%% file: main.tex
\documentclass[10pt,twocolumn,letterpaper]{article}

\usepackage{iccv}
\usepackage{times}
\usepackage{epsfig}
\usepackage{graphicx}
\usepackage{amsmath}
\usepackage{amssymb}
\usepackage{booktabs}
\usepackage{subcaption}

\usepackage[pagebackref=true,breaklinks=true,letterpaper=true,colorlinks,bookmarks=false]{hyperref}

\iccvfinalcopy 

\def\iccvPaperID{1532} 

\begin{document}

\title{Multi-Object Discovery by Low-Dimensional Object Motion}

\author{Sadra Safadoust \qquad Fatma Güney\\
KUIS AI Center and Department of Computer Engineering, Koç University\\
{\tt\small ssafadoust20@ku.edu.tr} \qquad {\tt\small fguney@ku.edu.tr}
}

\maketitle

\input{commands}
\input{sec/abstract}

\input{sec/intro}
\input{sec/rw}

\input{sec/method}

\input{sec/exp}

\input{sec/conclusion}

\boldparagraph{Acknowledgements}
Sadra Safadoust was supported by KUIS AI Fellowship and UNVEST R\&D Center.
\clearpage
{\small
\bibliographystyle{ieee_fullname}
\bibliography{bibliography_short, egbib}
}

\clearpage
\clearpage

\twocolumn[{%
  \begin{@twocolumnfalse}
    \supptitle{Supplementary Material for \\
``Multi-Object Discovery by Low-Dimensional Object Motion''}
  \end{@twocolumnfalse}
}]

\def\thesection {\Alph{section}}
\setcounter{section}{0}
\renewcommand{\theHsection}{Supplement.\thesection}

\input{sec/supp/summary}
\input{sec/supp/basis}

\input{sec/supp/projection}
\input{sec/supp/qual}
\input{sec/supp/movidepth}
\input{sec/supp/kitti-intrinsics}

\end{document}

%% file: commands.tex
\newcommand{\Perp}{\perp\!\!\! \perp}
\newcommand{\bK}{\mathbf{K}}
\newcommand{\bX}{\mathbf{X}}
\newcommand{\bY}{\mathbf{Y}}
\newcommand{\bk}{\mathbf{k}}
\newcommand{\bx}{\mathbf{x}}
\newcommand{\by}{\mathbf{y}}
\newcommand{\bhy}{\hat{\mathbf{y}}}
\newcommand{\bty}{\tilde{\mathbf{y}}}
\newcommand{\bG}{\mathbf{G}}
\newcommand{\bI}{\mathbf{I}}
\newcommand{\bg}{\mathbf{g}}
\newcommand{\bS}{\mathbf{S}}
\newcommand{\bs}{\mathbf{s}}
\newcommand{\bM}{\mathbf{M}}
\newcommand{\bw}{\mathbf{w}}
\newcommand{\eye}{\mathbf{I}}
\newcommand{\bU}{\mathbf{U}}
\newcommand{\bV}{\mathbf{V}}
\newcommand{\bW}{\mathbf{W}}
\newcommand{\bn}{\mathbf{n}}
\newcommand{\bv}{\mathbf{v}}
\newcommand{\bwv}{\mathbf{wv}}
\newcommand{\bq}{\mathbf{q}}
\newcommand{\bR}{\mathbf{R}}
\newcommand{\bi}{\mathbf{i}}
\newcommand{\bj}{\mathbf{j}}
\newcommand{\bp}{\mathbf{p}}
\newcommand{\bt}{\mathbf{t}}
\newcommand{\bJ}{\mathbf{J}}
\newcommand{\bu}{\mathbf{u}}
\newcommand{\bB}{\mathbf{B}}
\newcommand{\bD}{\mathbf{D}}
\newcommand{\bz}{\mathbf{z}}
\newcommand{\bP}{\mathbf{P}}
\newcommand{\bC}{\mathbf{C}}
\newcommand{\bA}{\mathbf{A}}
\newcommand{\bZ}{\mathbf{Z}}
\newcommand{\bff}{\mathbf{f}}
\newcommand{\bF}{\mathbf{F}}
\newcommand{\bo}{\mathbf{o}}
\newcommand{\bO}{\mathbf{O}}
\newcommand{\bc}{\mathbf{c}}
\newcommand{\bm}{\mathbf{m}}
\newcommand{\bT}{\mathbf{T}}
\newcommand{\bQ}{\mathbf{Q}}
\newcommand{\bL}{\mathbf{L}}
\newcommand{\bl}{\mathbf{l}}
\newcommand{\ba}{\mathbf{a}}
\newcommand{\bE}{\mathbf{E}}
\newcommand{\bH}{\mathbf{H}}
\newcommand{\bd}{\mathbf{d}}
\newcommand{\br}{\mathbf{r}}
\newcommand{\be}{\mathbf{e}}
\newcommand{\bb}{\mathbf{b}}
\newcommand{\bh}{\mathbf{h}}
\newcommand{\bhh}{\hat{\mathbf{h}}}
\newcommand{\btheta}{\boldsymbol{\theta}}
\newcommand{\bTheta}{\boldsymbol{\Theta}}
\newcommand{\bpi}{\boldsymbol{\pi}}
\newcommand{\bphi}{\boldsymbol{\phi}}
\newcommand{\bpsi}{\boldsymbol{\psi}}
\newcommand{\bPhi}{\boldsymbol{\Phi}}
\newcommand{\bmu}{\boldsymbol{\mu}}
\newcommand{\bsigma}{\boldsymbol{\sigma}}
\newcommand{\bSigma}{\boldsymbol{\Sigma}}
\newcommand{\bGamma}{\boldsymbol{\Gamma}}
\newcommand{\bbeta}{\boldsymbol{\beta}}
\newcommand{\bomega}{\boldsymbol{\omega}}
\newcommand{\blambda}{\boldsymbol{\lambda}}
\newcommand{\bLambda}{\boldsymbol{\Lambda}}
\newcommand{\bkappa}{\boldsymbol{\kappa}}
\newcommand{\btau}{\boldsymbol{\tau}}
\newcommand{\balpha}{\boldsymbol{\alpha}}
\newcommand{\nR}{\mathbb{R}}
\newcommand{\nN}{\mathbb{N}}
\newcommand{\nL}{\mathbb{L}}
\newcommand{\nE}{\mathbb{E}}
\newcommand{\cN}{\mathcal{N}}
\newcommand{\cM}{\mathcal{M}}
\newcommand{\cR}{\mathcal{R}}
\newcommand{\cB}{\mathcal{B}}
\newcommand{\cL}{\mathcal{L}}
\newcommand{\cH}{\mathcal{H}}
\newcommand{\cS}{\mathcal{S}}
\newcommand{\cT}{\mathcal{T}}
\newcommand{\cO}{\mathcal{O}}
\newcommand{\cC}{\mathcal{C}}
\newcommand{\cP}{\mathcal{P}}
\newcommand{\cE}{\mathcal{E}}
\newcommand{\cI}{\mathcal{I}}
\newcommand{\cF}{\mathcal{F}}
\newcommand{\cK}{\mathcal{K}}
\newcommand{\cY}{\mathcal{Y}}
\newcommand{\cX}{\mathcal{X}}
\def\bgamma{\boldsymbol\gamma}

\newcommand{\specialcell}[2][c]{%
  \begin{tabular}[#1]{@{}c@{}}#2\end{tabular}}

\newcommand{\figref}[1]{\Fig~\ref{#1}}
\newcommand{\secref}[1]{Section~\ref{#1}}
\newcommand{\algref}[1]{Algorithm~\ref{#1}}
\newcommand{\eqnref}[1]{Eq.~\eqref{#1}}
\newcommand{\tabref}[1]{Table~\ref{#1}}

\newcommand{\rulesep}{\unskip\ \vrule\ }


\newcommand{\KLD}[2]{D_{\mathrm{KL}} \Big(#1 \mid\mid #2 \Big)}

\renewcommand{\b}{\ensuremath{\mathbf}}

\def\mc{\mathcal}
\def\mb{\mathbf}

\newcommand{\T}{^{\raisemath{-1pt}{\mathsf{T}}}}

\makeatletter
\DeclareRobustCommand\onedot{\futurelet\@let@token\@onedot}
\def\@onedot{\ifx\@let@token.\else.\null\fi\xspace}
\def\eg{e.g\onedot} \def\Eg{E.g\onedot}
\def\ie{i.e\onedot} \def\Ie{I.e\onedot}
\def\cf{cf\onedot} \def\Cf{Cf\onedot}
\def\etc{etc\onedot} \def\vs{vs\onedot}
\def\wrt{wrt\onedot}
\def\dof{d.o.f\onedot}
\def\etal{et~al\onedot} \def\iid{i.i.d\onedot}
\def\Fig{Fig\onedot} \def\Eqn{Eqn\onedot} \def\Sec{Sec\onedot} \def\Alg{Alg\onedot}
\makeatother

\newcommand{\xdownarrow}[1]{%
  {\left\downarrow\vbox to #1{}\right.\kern-\nulldelimiterspace}
}

\newcommand{\xuparrow}[1]{%
  {\left\uparrow\vbox to #1{}\right.\kern-\nulldelimiterspace}
}


\newcommand*\rot{\rotatebox{90}}
\newcommand{\boldparagraph}[1]{\vspace{0.2cm}\noindent{\bf #1.} }
\newcommand{\boldquestion}[1]{\vspace{0.2cm}\noindent{\bf #1} }

\newcommand{\safa}[1]{ \noindent {\color{blue} {\bf Sadra:} {#1}} } 
\newcommand{\ftm}[1]{ \noindent {\color{magenta} {\bf Fatma:} {#1}} }

\newcommand{\supptitle}[1]{
   \newpage
   \null
  \vskip .375in
   \begin{center}
      {\Large \bf #1 \par}
      \vspace*{24pt}
      {
      \large
      \lineskip .5em
      \begin{tabular}[t]{c}
         Sadra Safadoust \qquad Fatma Güney\\
KUIS AI Center and Department of Computer Engineering, Koç University\\
{\tt\small ssafadoust20@ku.edu.tr} \qquad {\tt\small fguney@ku.edu.tr}\\
         \vspace*{1pt}\\
      \end{tabular}
      \par
      }
   \end{center}
   }

%% file: sec/abstract.tex
\begin{abstract}
Recent work in unsupervised multi-object segmentation shows impressive results by predicting motion from a single image despite the inherent ambiguity in predicting motion without the next image. On the other hand, the set of possible motions for an image can be constrained to a low-dimensional space
by considering the scene structure and moving objects in it.
We propose to model pixel-wise geometry and object motion to remove ambiguity in reconstructing flow from a single image.
Specifically, we divide the image into coherently moving regions and use depth to construct flow bases that best explain the observed flow in each region. 
We achieve state-of-the-art results in unsupervised multi-object segmentation on synthetic and real-world datasets by modeling the scene structure and object motion. 
Our evaluation of the predicted depth maps shows reliable performance in monocular depth estimation.
\end{abstract}

%% file: sec/intro.tex
\section{Introduction}
Finding objects on visual data is one of the oldest problems in computer vision, which has been shown to work to great extent in the presence of labeled data. Achieving it without supervision is important given the difficulty of obtaining pixel-precise masks for the variety of objects encountered in everyday life. In the absence of labels, motion provides important cues to group pixels corresponding to objects. The existing solutions use motion either as input to perform grouping or as output to reconstruct as a way of verifying the predicted grouping. The current methodology fails to incorporate the underlying 3D geometry creating the observed motion. In this work, we show that modeling geometry together with object motion significantly improves the segmentation of multiple objects without supervision.

Unsupervised multi-object discovery is significantly more challenging than the single-object case due to mutual occlusions. Therefore, earlier methods in unsupervised segmentation focused on separating a foreground object from the background whereas multi-object methods have been mostly limited to synthetic datasets or resorted to additional supervision on real-world data such as sparse depth~\cite{Elsayed2022NeurIPS}. 

While sparse-depth supervision can be applied to driving scenarios~\cite{Elsayed2022NeurIPS}, depth information is not typically available on common video datasets. Moreover, video segmentation datasets such as DAVIS~\cite{Perazzi2016CVPR, Pont-Tuset2017ARXIV} contain a wide variety of categories under challenging conditions such as appearance changes due to lighting conditions or motion blur. The motion information can be obtained from video sequences via optical flow. Optical flow not only provides motion cues for grouping~\cite{Yang2021ICCV} but can also be used for training on synthetic data without suffering from the domain gap while transferring to real data~\cite{Xie2022NeurIPS}. The problems in optical flow prediction on real-world data can be mitigated to some extent by relating flow predictions from multiple frames~\cite{Xie2022NeurIPS}.

In addition to problems in predicting optical flow, requiring flow as input prohibits the application of the method on static images. Another line of work~\cite{Choudhury2022BMVC, Karazija2022NeurIPS} uses motion for supervision at train time only. Based on the observation that objects create distinctive patterns in flow, initial work~\cite{Choudhury2022BMVC} fits a simple parametric model to the flow in each object region to capture the object motion. This way, the network can predict object regions that can potentially move coherently from a single image at test time. There is an inherent ambiguity in predicting motion from a single image. Therefore, the follow-up work~\cite{Karazija2022NeurIPS} predicts a distribution of possible motion patterns to reduce this ambiguity. This also allows extending it to the multi-object case by mitigating the over-segmentation problem of the initial work~\cite{Choudhury2022BMVC}.

In this work, we propose to model pixel-wise geometry to remove ambiguity in reconstructing flow from a single image. Optical flow is the difference between the 2D projections of the 3D world in consecutive time steps. By modeling the 3D geometry creating these projections, we directly address the mutual occlusion problem due to interactions of multiple objects. This problem has been crudely addressed by previous work with a depth-ordered layer representation~\cite{Xie2022NeurIPS}. Instead of assuming a single depth layer per object, we predict pixel-wise depth which provides more expressive power in explaining the observed motion. Furthermore, we do not use flow as input during inference, allowing us to evaluate our method on single-image datasets. 

Recent work~\cite{Bowen2022THREEDV} showed that motion resides in a low-dimensional subspace, and its reconstruction can be used to supervise monocular depth prediction. Despite many possible flow fields, the space of possible flow fields is spanned by a small number of basis flow fields related to depth and independently moving objects. While \cite{Bowen2022THREEDV} focuses on modeling camera motion for quantitatively evaluating depth in static scenes, it also points to the fact that the object motion can be similarly modeled in a low-dimensional subspace by simply masking the points in the object region. Given the difficulty of predicting pixel-wise masks, simple object embeddings are used to cluster independently moving objects. We instead predict the object regions jointly with depth to find the low-dimensional object motion that best explains the observed flow in each region.

Our approach works extremely well on synthetic Multi-Object Video (MOVi) datasets~\cite{Greff2022CVPR}, significantly outperforming previous work, especially in more challenging MOVi-\{C,D,E\} partitions and performing comparably on visually simpler MOVi-A due to difficulty of estimating depth.
We use motion only for supervision at train time, therefore our method can be successfully applied to still images of CLEVR~\cite{Johnson2017CVPR} and C{\footnotesize LEVR}T{\footnotesize EX}~\cite{Karazija2021NeurIPS} and shows state-of-the-art performance.
More impressively, our method can segment multiple objects on real-world videos of DAVIS-2017~\cite{Pont-Tuset2017ARXIV} from a single image at test time, exceeding the performance of the state-of-the-art that uses flow from multiple frames as input~\cite{Xie2022NeurIPS}. 
In addition to evaluating segmentation, we show that our method can also reliably predict depth in real-world self-driving
scenarios on KITTI~\cite{Geiger2012CVPR}.

%% file: sec/rw.tex
\section{Related Work}
\boldparagraph{Basis Learning} Early work showed that optical flow estimation due to camera motion can be constrained using a subspace formulation for flow \cite{IraniICCV1999}. Basis learning has been used as a regularization in low-level vision, unifying tasks such as depth, flow, and segmentation~\cite{Tang2020CVPR}. 
PCA-Flow~\cite{Wulff2015CVPR} builds a higher dimensional flow subspace from movies to represent flow as a weighted sum of flow bases.
Recent work \cite{Ye2021ICCV} learns the coefficients to combine eight pre-defined flow bases for homography estimation.

\boldparagraph{Motion as Input} Most of the work in motion segmentation focuses on the single-object case. While earlier work uses traditional methods to cluster pixels into similar motion groups~\cite{Brox2010ECCV, Keuper2015ICCV, Ochs2011ICCV, Xie2019CVPR}, later methods train deep neural networks which take flow as input and predict segmentation as output~\cite{Dave2019ICCVW, Tokmakov2017CVPR, Tokmakov2019IJCV}. Another work~\cite{Yang2019CVPR} uses the distinctiveness of motion in the case of foreground objects by proposing an adversarial setting to predict motion from context. 
Segmenting objects in camouflaged settings can be achieved by modeling background motion to remove its effect and highlight the moving foreground object~\cite{Bideau2016ECCV, Bideau2018CVPR, Lamdouar2020ACCV}.
Recent work uses consistency between two flow fields computed under different frame gaps for self-supervision~\cite{Yang2021ICCV}.

The most relevant to our work is OCLR~\cite{Xie2022NeurIPS} which extends motion segmentation to multiple objects by relating motion extracted from multiple frames using a transformer in a layered representation. In this work, we show that better results can be achieved on real data even from a single image by modeling pixel-wise geometry. 

\boldparagraph{Motion for Supervision}
While using motion only as input works well where appearance fails, \eg the camouflage datasets, RGB carries important information that might be missing in flow.
DyStaB~\cite{Yang2021CVPR} trains a dynamic model by exploiting motion for temporal consistency and then uses it to bootstrap a static model which takes a single image as input.
A single image network is used to predict a segmentation in \cite{Liu2021NeurIPS} and then the motion of each segment is predicted with a two-frame motion network.
While image warping loss is used in \cite{Liu2021NeurIPS} for self-supervision, recent work~\cite{Choudhury2022BMVC, Karazija2022NeurIPS} uses flow reconstruction loss by assuming the availability of flow at train time only.
GWM~\cite{Choudhury2022BMVC} segments foreground objects by fitting an approximate motion model to each segment and then merging them using spectral clustering.
The follow-up work~\cite{Karazija2022NeurIPS} extends it to multiple objects by predicting probable motion patterns for each segment with a distribution. We also reconstruct flow for supervision but differently, we account for 3D to remove the ambiguity in reconstructing motion from a single image.

The most relevant to our work is the previous work that uses flow as a source of supervision for depth~\cite{Bowen2022THREEDV} or segmentation~\cite{Choudhury2022BMVC, Karazija2021NeurIPS}. In this work, we model both depth and segmentation with supervision from motion.

\boldparagraph{Multi-Object Scene Decomposition}
Our work is also related to scene decomposition approaches which are mostly evaluated on synthetic datasets. The earlier image-based decomposition approaches such as MONet~\cite{Burgess2019ARXIV} and IODINE~\cite{Greff2019ICML} use a sequential VAE structure where the decomposition at a step can affect the remaining parts to be explained in the next step. GENESIS~\cite{Engelcke2020ICLR} follows an object-centric approach by accounting for component interactions, which is extended to more realistic scenarios with an autoregressive prior in the follow-up work~\cite{EngelckeNeurIPS2021}. Slot Attention~\cite{Locatello2020NeurIPS} uses an iterative attention mechanism  to decompose the image into a set of slot representations.
A hierarchical VAE is used in \cite{Emami2021ICML} to extract symmetric and disentangled representations.

\input{figs/main_figure}

There are also video-based approaches to multi-object scene decomposition.
SCALOR~\cite{Jiang2020ICLR} focuses on scaling generative approaches to crowded scenes in terms of object density.
SIMONe~\cite{Kabra2021NeurIPS} learns a factorized latent space to separate object semantics that is constant in the sequence from the background which changes at each frame according to camera motion.
SAVi~\cite{Kipf2022ICLR} extends Slot Attention~\cite{Locatello2020NeurIPS} to videos and SAVi++~\cite{Elsayed2022NeurIPS} extends it to real-world driving scenarios with sparse depth supervision.

\boldparagraph{Self-Supervised Monocular Depth Estimation}
Zhou \etal \cite{Zhou2017CVPR} train a pose network to estimate the pose between the frames in a sequence and jointly train it with the depth network. Godard \etal \cite{Godard2019ICCV} improves the results with a better loss function and other design choices. Guizilini \etal \cite{Guizilini2020CVPR} learn detail-preserving representations using 3D packing and unpacking blocks.
Given instance segmentation masks, a line of work~\cite{Casser2019AAAI, Lee2021AAAI} models the motion of objects in the scene in addition to the camera motion to go beyond the static-scene assumption. While the object masks are supervised using ground truth masks in \cite{Klingner2020ECCV}, the masks are learned without supervision as an auxiliary output in \cite{Safadoust2021THREEDV} for better depth estimation. 
While they require multiple frames during inference, our approach can estimate masks from a single image. Additionally, our method does not use camera intrinsics.

%% file: figs/main_figure.tex
\begin{figure*}[t]
    \centering
    \includegraphics[width=\linewidth]{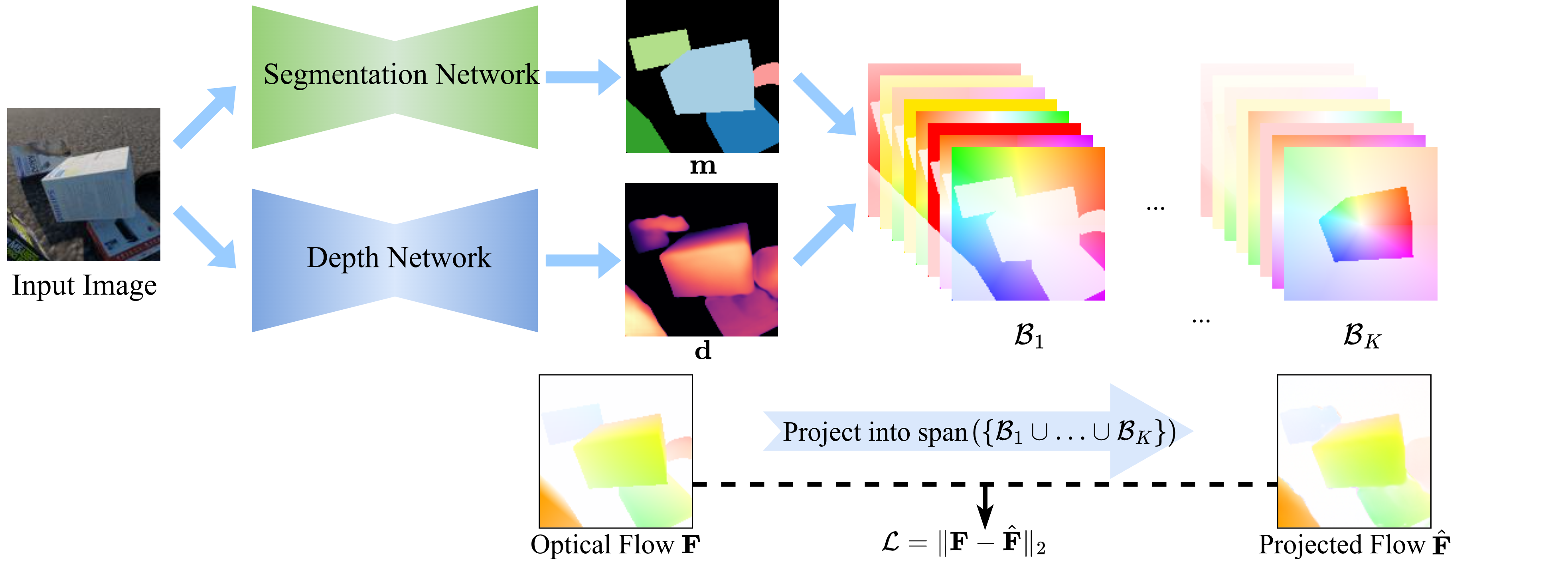}
    \caption{\textbf{Overview of our Approach.} From a single image, we use a segmentation and a depth network to predict a segmentation mask $\bm$ and a disparity map $\bd$. Based on these predictions, we construct the bases for the space of the possible optical flows for $K$ distinctly moving regions on the image. Each moving region $i$ is represented with a separate basis $\mathcal{B}_i$. Given optical flow $\mathbf{F}$ as input, either ground truth or estimated by an off-the-shelf method, we project it into span$(\bigcup_{i=1}^K\mathcal{B}_i)$. We use the distance between the input flow $\mathbf{F}$ and the projected flow $\hat{\mathbf{F}}$ to supervise depth and segmentation. During inference, our networks can be used to predict depth and segmentation from a single image. 
    }
    \label{fig:main}
\end{figure*}

%% file: sec/method.tex
\section{Depth-Aware Multi-Object Segmentation}
\label{sec:method}
The observed motion in 2D is the result of 3D scene structure and independently moving objects.
By predicting the scene structure in terms of depth and locating independently moving objects, we can accurately reconstruct the optical flow corresponding to the observed motion in 2D.
Towards this purpose, we use a low-dimensional parameterization of optical flow based on depth (\secref{sec:low-dim-motion}). In this low-dimensional representation, we can accurately reconstruct flow from a rigid motion. We extend this parametrization to a number of rigidly moving objects to find the regions corresponding to objects (\secref{sec:object_motion}).
See \figref{fig:main} for an overview of our approach.

\subsection{Low-Dimensional Motion Representation}
\label{sec:low-dim-motion}
The space of all possible optical flow fields is very high-dimensional, \ie in $\nR^{H \times W \times 2}$. However, conditioned on the scene structure, only a small fraction of all flow fields are possible.
Previous work ~\cite{Heeger1992IJCV, Bowen2022THREEDV} has shown that the set of possible instantaneous flows for a moving camera in a static scene forms a linear space with six basis vectors:
\begin{equation}
    \label{eq:six_basis}
    \mathcal{B}_0 = \{\bb_{\bT x}, \bb_{\bT y}, \bb_{\bT z}, \bb_{\bR x}, \bb_{\bR y}, \bb_{\bR z}\}
\end{equation}
These basis vectors correspond to translation and rotation along the $x, y$, and $z$ axes, respectively. For an image $\bI \in \nR^{H\times W \times 3}$, the values of each basis vector $\bb_i \in \nR^{H \times W \times 2}$ for a given pixel $(u, v)$ can be calculated as follows:
\begin{eqnarray}
\label{eqn:basis-camera}
        &\bb_{\bT x} = \begin{bmatrix}
              f_x~d \\
             0
          \end{bmatrix}, 
        &\bb_{\bR x} =  \begin{bmatrix}
              {f_y}^{-1}~\bar{u}~\bar{v} \\
             f_y + {f_y}^{-1}~\bar{v}^2
          \end{bmatrix} \nonumber \\
        \nonumber \\ %
        &\bb_{\bT y} = \begin{bmatrix}
              0 \\
             f_y~d
          \end{bmatrix}, 
        &\bb_{\bR y} = \begin{bmatrix}
            f_x + {f_x}^{-1}~\bar{u}^2 \\
            {f_x}^{-1}~\bar{u}~\bar{v}
        \end{bmatrix} \nonumber \\
        \nonumber \\ %
        &\bb_{\bT z} = \begin{bmatrix}
             -\bar{u}~d \\
             -\bar{v}~d
          \end{bmatrix},
        &\bb_{\bR z} = \begin{bmatrix}
             f_x~{f_y}^{-1}~\bar{v} \\
             -f_y~{f_x}^{-1}~\bar{u}
          \end{bmatrix}  
\end{eqnarray}
%
where $f_x, f_y$ are the focal lengths of the camera. For brevity, we define $\bar{u} = u - c_x$ and $\bar{v} = v - c_y$ to be the centered pixel coordinates according to $(c_x, c_y)$, the principal point of the camera.
With a slight abuse of notation, we do not write the basis vectors as a function of $(u,v)$ and use $d$ to denote the disparity $\bd(u, v)$ at a pixel $(u, v)$. 

We train a monocular depth network to predict inverse depth, disparity $\bd$ from a single image. Then, the predicted disparity at each pixel is used to form the translation part of the basis vectors as shown in \eqnref{eqn:basis-camera}. Note that predicted disparity values do not affect the rotation but form the low-dimensional motion representation via translation. The depth network receives gradients directly from the flow reconstruction loss as explained next in \secref{sec:object_motion}.
 
In \eqnref{eqn:basis-camera}, camera parameters including the principal point $(c_x, c_y)$ and the focal lengths $f_x$, $f_y$ are required to calculate the basis vectors. We assume the principal points to be at the center of the image. However, we do not assume to know the values of focal lengths. Instead, we only assume that $f_x = f_y$. In this case, as demonstrated by \cite{Bowen2022THREEDV}, we can rewrite $\mathcal{B}_0$ as a set of $8$ vectors that do not depend on the values of $f_x$ and $f_y$. For more details, please see Supplementary.
 Even without knowing the actual values of the focal lengths, we can obtain quite accurate depth predictions with supervision from flow (\secref{sec:exp}).



\subsection{Segmentation by Object Motion}
\label{sec:object_motion}
We extend the formulation introduced in \secref{sec:low-dim-motion} to handle the instantaneous flow from multiple object motion. As stated in \cite{Bowen2022THREEDV}, for a rigidly moving object in the scene, there is an equivalent camera motion. Therefore the space of optical flow from a rigidly moving object is the same as the space of optical flow from camera motion restricted to points in the object. Consider a scene with $K$ regions corresponding to moving parts including the background and multiple objects. If we represent each region $i \in \{1, \dots, K\}$ with ones on a binary mask $\bm_i \in \{0, 1\} ^ {H \times W \times 1}$, then a basis for the space of possible flows can be defined as follows:
\begin{equation}
\label{eqn:basis-all}
    \mathcal{B} = \{\mathcal{B}_1 \cup \mathcal{B}_2 \cup \ldots \mathcal{B}_K \} 
\end{equation}
where $\mathcal{B}_i$ refers to the basis for the space of possible flows  restricted to region $i$ as:
\begin{equation}
\label{eqn:basis-region}
\mathcal{B}_i = \{ \bm_i \bb \mid \bb \in \mathcal{B}_0 \}.
\end{equation}
We train a segmentation network to divide the image into coherently moving regions $\bm \in [0, 1 ]^{H \times W \times K}$, representing soft assignments over $K$ regions. We use the predicted mask $\bm_i \in [0, 1 ]^{H \times W \times 1}$ of the region $i$ to obtain the basis corresponding to that region according to \eqnref{eqn:basis-region}.
%
%

\input{figs/movi}


\boldparagraph{Training Objective}
Based on the predicted disparity map $\bd$ and the segmentation map $\bm$, we form the basis $\mathcal{B}$ for the space of possible optical flows for the image according to \eqnref{eqn:basis-all} and \eqnref{eqn:basis-region}. 
We denote the optical flow where the input image is the source frame as $\bF \in \nR^{H \times W \times 2}$. It can be either ground truth flow or the output of a two-frame flow network such as RAFT~\cite{Teed2020ECCV}. 
We project $\bF$ into the space spanned by $\mathcal{B}$ in a differentiable manner to obtain $\hat{\bF}$. For the details of the projection, please refer to Supplementary. We define our loss function as the $L_2$ distance between the given flow $\bF$ and the reconstructed flow $\hat{\bF}$ and use it to train depth and segmentation networks jointly:
\begin{equation}
\cL = \lVert \bF - \hat{\bF} \rVert_2
\end{equation}
%

%% file: figs/movi.tex
\begin{figure*}[t]
    \centering
 \rotatebox[origin=l]{90}{
 \parbox{0.09\linewidth}{ Our depth}
 \parbox{0.09\linewidth}{\centering Ours}
 \parbox{0.09\linewidth}{\centering PPMP}
 \parbox{0.09\linewidth}{\centering GT Seg}
 \parbox{0.09\linewidth}{\centering Flow}
 \parbox{0.09\linewidth}{\centering RGB}}
 \hspace{0.001\linewidth}
 \includegraphics[width=0.093\linewidth]{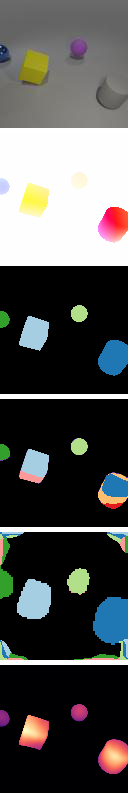}
    \includegraphics[width=0.093\linewidth]{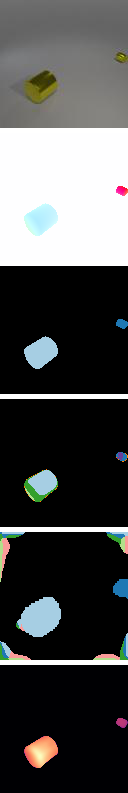}
    \hspace{0.002\linewidth}
    \includegraphics[width=0.093\linewidth]{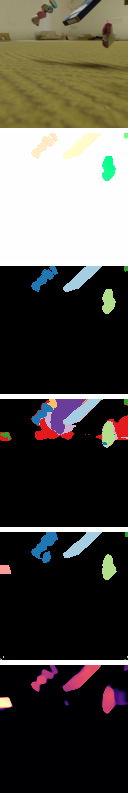}
    \includegraphics[width=0.093\linewidth]{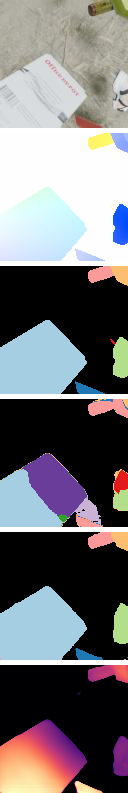}
    \hspace{0.002\linewidth}
    \includegraphics[width=0.093\linewidth]{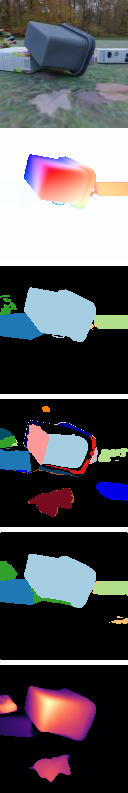}
    \includegraphics[width=0.093\linewidth]{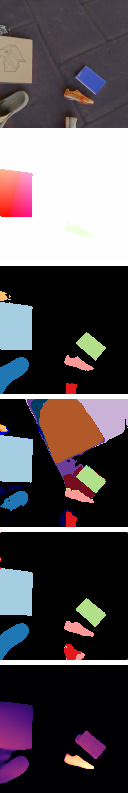}
    \hspace{0.002\linewidth}
    \includegraphics[width=0.093\linewidth]{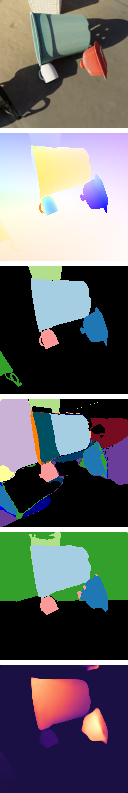}
    \includegraphics[width=0.093\linewidth]{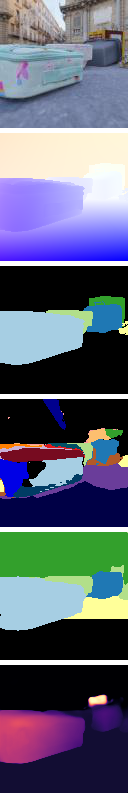}
    \\
     \quad MOVi-A  \quad \quad  \quad  \quad  \quad \quad MOVi-C \quad \quad \quad \quad  \quad  \quad  MOVi-D \quad \quad \quad \quad  \quad  \quad  \quad MOVi-E

    \caption{\textbf{Visualization of Depth and Segmentation Results on MOVi datasets}. Our method performs accurate segmentations, while PPMP suffers from over-segmentation and also mistakenly segments parts of the background as objects .}
    \label{fig:movi}
    \vspace{-0.3cm}
\end{figure*}

%% file: sec/exp.tex
\section{Experiments}
\label{sec:exp}
%

\subsection{Datasets}
\label{sec:datasets}
\boldparagraph{Synthetic Datasets} For comparison to image-based methods, we evaluate our method on the CLEVR \cite{Johnson2017CVPR} and C{\footnotesize LEVR}T{\footnotesize EX} \cite{Karazija2021NeurIPS} datasets. CLEVR is a dataset of still images depicting multiple objects of random shape, size, color, and position. C{\footnotesize LEVR}T{\footnotesize EX} is similar to CLEVR but contains more diverse textures and shapes. Because our method needs optical flow for training, we train our model on the M{\footnotesize OVING}CLEVR and M{\footnotesize OVING}C{\footnotesize LEVR}T{\footnotesize EX} datasets~\cite{Karazija2022NeurIPS}, which are video extensions of CLEVR and C{\footnotesize LEVR}T{\footnotesize EX} scenes. We train on the video versions but evaluate on the original test sets of CLEVR and C{\footnotesize LEVR}T{\footnotesize EX}. 

For comparison to video-based methods, we use the Multi-Object Video (MOVi) datasets~\cite{Greff2022CVPR}. Similar to~\cite{Karazija2022NeurIPS}, we use the MOVi-\{A, C, D, E\} variants. MOVi-A is based on CLEVR~\cite{Johnson2017CVPR} and contains scenes with a static camera and multiple objects with simple textures and uniform colors tossed on a gray floor. MOVi-C is more challenging due to realistic everyday objects with rich textures on a more complex background. MOVi-D increases the complexity by increasing the number of objects. MOVi-E is even more challenging as it features camera motion as well.

In all our experiments on synthetic datasets, we use a resolution of $128\times 128$ and the ground truth optical flow.

\boldparagraph{Real-World Datasets} We use the common video segmentation dataset DAVIS-2017~\cite{Pont-Tuset2017ARXIV} containing 90 video sequences where each sequence has one or more moving objects. We follow the evaluation protocol of~\cite{Xie2022NeurIPS} where the ground truth objects are reannotated by assigning the same label to the jointly moving objects. We resize the images to a resolution of $128 \times 224$ during training and use the flow from RAFT~\cite{Teed2020ECCV} with $\{ -8, -4, 4, 8 \}$ gaps between frames. 

Additionally, we evaluate our method on the KITTI driving dataset~\cite{Geiger2012CVPR, Geiger2013IJRR}. Following~\cite{Bao2022CVPR}, we train on the whole training set and evaluate the segmentation results on the instance segmentation benchmark consisting of 200 frames. We use a resolution of $128 \times 416$ and the flow from RAFT~\cite{Teed2020ECCV} with a gap of $+1$. Additionally, we evaluate our depth results on KITTI. Following prior work~\cite{Zhou2017CVPR, Godard2019ICCV}, we evaluate depth on the Eigen split~\cite{Eigen2014NeurIPS} of the KITTI dataset using improved ground truth~\cite{Uhrig2017THREEDV} to be comparable to self-supervised monocular depth estimation approaches. 

\subsection{Architecture Details}
We use the same architecture used in~\cite{Ranftl2020TPAMI} for depth and Mask2Former~\cite{Cheng2022CVPR} for segmentation, using only the segmentation head. 
We use different backbones for the segmentation network on the synthetic and real datasets.
On synthetic datasets, we follow \cite{Karazija2022NeurIPS, Kipf2022ICLR, Locatello2020NeurIPS} and utilize a 6-layer CNN.
On real-world datasets, following~\cite{Choudhury2022BMVC}, we use a ViT-B/8 transformer pre-trained self-supervised using DINO~\cite{Caron2021ICCV} on ImageNet~\cite{Russakovsky2015IJCV}. On all of the datasets, we use $6$ object queries in the segmentation network, which translates to $K=6$, except for CLEVR, where we use $K=8$. 

We use a fixed learning rate of $5 \times 10^{-5}$ for the depth network and use $1.5 \times 10^{-4}$ with a linear warm-up for the first 5K iterations for the segmentation network, reduced to $1.5 \times 10^{-5}$ after 200K iterations. We train both networks with AdamW optimizer~\cite{loshchilov2017ARXIV} for 250K iterations. See Supplementary for further details, we will also share the code.

\subsection{Evaluation Details}
\boldparagraph{Metrics} Following prior work \cite{Karazija2021NeurIPS, Kipf2022ICLR, Karazija2022NeurIPS}, we evaluate segmentation on synthetic datasets using Adjusted Rand Index on foreground pixels (FG-ARI) and mean Intersection over Union (mIoU). ARI 
measures how well predicted and ground truth segmentation masks match in a permutation-invariant manner. For mIoU, we first apply Hungarian matching and calculate the mean over the maximum between the number of ground-truth and predicted segments. 

\input{tab/movi}

On DAVIS-2017~\cite{Pont-Tuset2017ARXIV}, we use the standard $\mathcal{J}$, $\mathcal{F}$ metrics and perform the Hungarian matching per frame, similar to other datasets. Note that we focus on the multi-object segmentation task without using any labels for segmentation at train or test time. 
For KITTI, we use the FG-ARI metric, following \cite{Karazija2022NeurIPS, Bao2022CVPR}. 
For evaluating depth, we use the standard metrics used in monocular depth estimation~\cite{Eigen2014NeurIPS, Garg2016ECCV}.

\boldparagraph{Post-processing} We also report the results on segmentation using the post-processing method introduced in \cite{Karazija2022NeurIPS}. They extract the connected components in the model output and choose the largest $K$ masks and discard any masks that take up less than $0.1\%$ of the image. Then they combine the discarded masks with the largest mask. The results with post-processing are marked with $^\dag$ in the tables.

\subsection{Results on Synthetic Datasets}
We evaluate our method on synthetic datasets and compare its performance to both image-based and video-based methods. Our method uses motion during training only. Therefore, it can also be evaluated on the image datasets.

\boldparagraph{Video-Based Methods} We compare our method to other video-based methods on the MOVi video datasets in \tabref{tab:movi} and \figref{fig:movi}. All of the methods in \tabref{tab:movi} use optical flow for supervision. Differently, SCALOR~\cite{Jiang2020ICLR} and SAVi~\cite{Kipf2022ICLR} use all frames in a video, whereas PPMP~\cite{Karazija2022NeurIPS} and our method perform single-image segmentation, one frame at a time without using any motion information at test time. 
On the simpler MOVi-A dataset, the performance of our method falls behind SAVi~\cite{Kipf2022ICLR} and PPMP~\cite{Karazija2022NeurIPS}.
PPMP with Swin transformer~\cite{Liu2021ICCV} performs the best overall. With the same 6-layer CNN backbone and without post-processing, SAVi performs the best. Despite the advantage of motion information, the success of SCALOR~\cite{Jiang2020ICLR} and SAVi~\cite{Kipf2022ICLR} is limited to visually simpler MOVi-A.

On the more challenging MOVi-\{C,D,E\} datasets, our method, even without post-processing, significantly outperforms all the previous methods in both metrics, with or without post-processing. 
The previous state-of-the-art, PPMP~\cite{Karazija2022NeurIPS} uses the same backbone in their segmentation network as ours. Even with a more powerful backbone (Swin transformer~\cite{Liu2021ICCV}) and post-processing, the results of PPMP are still far behind our results without any post-processing. 
From MOVi-C to MOVi-E, the performance gap between our method and the others increases as the complexity of the dataset increases. 
Please see Supplementary for qualitative results with post-processing and evaluation of our estimated depth for objects in these datasets.
%
\input{tab/clevr_long}

\boldparagraph{Image-Based Methods} We compare our method to other image-based methods on the CLEVR and C{\footnotesize LEVR}T{\footnotesize EX} datasets in \tabref{tab:clevr-long}. Our method outperforms the state-of-the-art method PPMP~\cite{Karazija2022NeurIPS} in all metrics on both datasets except for mIoU on CLEVR without postprocessing and mIoU on C{\footnotesize LEVR}T{\footnotesize EX} with postprocessing. We point to \textbf{+9.01} improvement in mIoU on the more challenging C{\footnotesize LEVR}T{\footnotesize EX} dataset without postprocessing. See Supplementary for qualitative results on CLEVR and C{\footnotesize LEVR}T{\footnotesize EX}.

\input{figs/davis17}

\subsection{Results on Real-World Datasets}
We compare our method to multi-object segmentation methods on real-world datasets including driving scenarios on KITTI and unconstrained videos on DAVIS-2017.

\input{tab/davis17}
\input{tab/kitti_combined}

\boldparagraph{Results on DAVIS} 
Our method is the first image-based method to report performance in multi-object segmentation without using any labels during training or testing on DAVIS-2017. Therefore, in \tabref{tab:davis}, we compare it to video-based approaches which use motion as input. We also compare to a simple baseline proposed in~\cite{Xie2022NeurIPS} based on Mask R-CNN~\cite{He2017ICCV} using optical flow as input.
We use the labels re-annotated by \cite{Xie2022NeurIPS} for evaluation, as explained in \secref{sec:datasets}. Motion Grouping refers to \cite{Yang2021ICCV} trained on DAVIS-2017. Motion Grouping (sup.), Mask R-CNN (flow) and OCLR are models trained on synthetic data from \cite{Xie2022NeurIPS} in a supervised way using optical flow as input. Our method, which uses a single RGB image as input at test time, outperforms previous methods, including the state-of-the-art OCLR~\cite{Xie2022NeurIPS} that uses flow from multiple time steps. Ours-M refers to a version of our model where we adapt the spectral clustering approach of \cite{Choudhury2022BMVC} to merge our predicted regions into the ground truth number of regions in each frame. Although not necessary to achieve state-of-the-art, this improves the results significantly. 

We visualize the results of our model in comparison to OCLR~\cite{Xie2022NeurIPS} in \figref{fig:davis17}. Our method can correctly segment a wide variety of objects such as the bike and the person in the first column and the multiple fish in the third, multiple people walking or fighting in the second and fourth. 
OCLR is highly affected by the inaccuracies in flow, unlike our method, as can be seen from the last two columns. 

%
\input{tab/ablation_movi}
\boldparagraph{Results on KITTI} 
Since KITTI has ground truth depth, we evaluate our method in terms of both segmentation and monocular depth prediction on KITTI.
The segmentation results on KITTI are presented in \tabref{tab:seg-kitti}. Our method is among the top-performing methods, outperforming earlier approaches. 
In addition to segmentation, our method can also predict depth from a single image.
We present the evaluation of our depth predictions in comparison to self-supervised monocular depth estimation methods in \tabref{tab:depth-kitti}. Our depth network can predict reliable depth maps even without camera intrinsics with comparable results to recent self-supervised monocular depth estimation approaches~\cite{Godard2019ICCV, Guizilini2020CVPR} that are specifically designed for that task and that use camera intrinsics.

We visualize our segmentation and depth results on KITTI in 
Supplementary. Our method can segment multiple moving objects such as car, bike, bus without using any semantic labels for training or motion information at test time. Furthermore, it can predict high-quality depth, capturing thin structures and sharp boundaries around objects. 

\subsection{Ablation Study}
To evaluate the contribution of different types of flow basis, we perform an experiment by considering only translation or rotation and compare it to the full model on MOVi datasets in \tabref{tab:movi_ablation}. Note that in the rotation-only case (Only-R), the depth predictions are not used and the depth network is not trained. 
Overall, the rotation-only model outperforms the translation-only model and the full model with both rotation and translation works the best on MOVi-\{C, D, E\} datasets with reliable depth predictions. 

The trend is different on the simpler MOVi-A dataset. Only-R outperforms all models including the state-of-the-art in \tabref{tab:movi}. We found that in the translation-only case, the depth cannot be predicted on MOVi-A due to missing texture and enough detail for the depth network to learn a mapping from a single image to depth. The rotation-only model, on the other hand, learns to group pixels in a region, based on their rotational motion which does not depend on depth. 
This ability explains the success of Only-R on simpler datasets. The importance of pixel-wise depth increases with the complexity of the dataset. On MOVi-E, for example, which has the most complex setup with a large number of objects and camera motion, predicting depth, from Only-R to Full,  improves the performance the most.




%% file: tab/movi.tex
\begin{table*}[t!]
\centering
\begin{tabular}{@{}lcccccccc@{}}
\toprule
  & \multicolumn{2}{c}{MOVi-A}    & \multicolumn{2}{c}{MOVi-C}    & \multicolumn{2}{c}{MOVi-D}    & \multicolumn{2}{c}{MOVi-E}      \\ \cmidrule(lr){2-3} \cmidrule(lr){4-5} \cmidrule(lr){6-7}\cmidrule(l){8-9}
  & \textbf{FG-ARI}$\uparrow$  & \textbf{mIoU}$\uparrow$  & \textbf{FG-ARI}$\uparrow$  & \textbf{mIoU}$\uparrow$  & \textbf{FG-ARI}$\uparrow$  & \textbf{mIoU}$\uparrow$  & \textbf{FG-ARI}$\uparrow$  & \textbf{mIoU}$\uparrow$  \\ \midrule 
GWM~\cite{Cheng2022CVPR} & 70.30 & 42.27 & 49.98 & 30.17 & 39.78 & 18.38 & 42.50 & 18.74 \\
SCALOR~\cite{Jiang2020ICLR}   & 59.57 & 44.41 & 40.43 & 22.54 & -     & -     & -     & -     \\
SAVi~\cite{Kipf2022ICLR} & \underline{88.30} & 62.69 & 43.26 & 31.92 & 43.45 & 10.60 & 17.39 & 5.75  \\
PPMP~\cite{Karazija2022NeurIPS} & 84.01 & 60.08 & 61.18 & 34.72 & 55.74 & 23.50 & 62.62 & 25.78 \\
PPMP$^\dag$ & 85.41 & \underline{76.19} & 61.24 & 37.26 & 55.18 & 25.21 & 63.11 & 28.59 \\
PPMP$^\dag$ (Swin) & \textbf{90.08} & \textbf{84.76} & 67.67 & 52.17 & 66.41 & 30.40 & 72.73 & 35.30 \\

Ours     & 56.09 & 36.48 & \underline{73.80} & \underline{54.48} & \underline{76.41} & \underline{58.82} & \underline{78.33} & \underline{47.38} \\
Ours$^\dag$  & 70.15 & 46.26 & \textbf{74.64} & \textbf{59.24} & \textbf{77.15} & \textbf{59.68} & \textbf{80.83} & \textbf{50.48} \\
\bottomrule
\end{tabular}
\caption{\textbf{Segmentation Results on MOVi Datasets.} The best result in each column is shown in \textbf{bold}, and the second best is \underline{underlined}. $^\dag$ indicates post-processing, and (Swin) denotes using a Swin transformer as the backbone.}
\label{tab:movi}
\vspace{-0.2cm}
\end{table*}

%% file: tab/clevr_long.tex
\begin{table}[b]
\centering
\begin{tabular}{@{}lcccc@{}}
\toprule
      & \multicolumn{2}{c}{CLEVR}      & \multicolumn{2}{c}{C{\footnotesize LEVR}T{\footnotesize EX} }    \\ \cmidrule(lr){2-3} \cmidrule(lr){4-5}
 & \textbf{FG-ARI}$\uparrow$ & \textbf{mIoU}$\uparrow$ & \textbf{FG-ARI}$\uparrow$ & \textbf{mIoU}$\uparrow$ \\ \midrule
SPAIR~\cite{Crawford2019AAAI} & 77.13 & \underline{65.95} & 0.00 & 0.00    \\
MN~\cite{Smirnov2021NeurIPS} & 72.12 & 56.81 & 38.31 & 10.46   \\
MONet~\cite{Burgess2019ARXIV} & 54.47 & 30.66 & 36.66 & 19.78   \\
SA~\cite{Locatello2020NeurIPS} & 95.89 & 36.61 & 62.40 & 22.58   \\
IODINE~\cite{Greff2019ICML} & 93.81 & 45.14 & 59.52 & 29.17   \\
DTI-S~\cite{Monnier2021ICCV} & 59.54 & 48.74 & 79.90 & 33.79   \\
GNM~\cite{Jiang2020NeurIPS} & 65.05 & 59.92 & 53.37 & 42.25   \\
SAVi~\cite{Kipf2022ICLR} & -    & -  & 49.54 & 31.88   \\
PPMP~\cite{Kabra2021NeurIPS} & 91.69 & \textbf{66.70} & \underline{90.80} & \underline{55.07}   \\
Ours  & \textbf{95.03} & 63.36 & \textbf{94.66} & \textbf{64.08}   \\
\midrule
SPAIR$^\dag$~\cite{Crawford2019AAAI} & 77.05 & 66.87 & 0.00 & 0.00    \\
MN$^\dag$~\cite{Smirnov2021NeurIPS} & 72.08 & 57.61 & 38.34 & 10.34   \\
MONet$^\dag$~\cite{Burgess2019ARXIV} & 61.36 & 45.61 & 35.64 & 23.59   \\
SA$^\dag$~\cite{Locatello2020NeurIPS} & 94.88 & 37.68 & 61.60 & 21.96   \\
IODINE$^\dag$~\cite{Greff2019ICML} & 93.68 & 44.20 & 60.63 & 29.40   \\
DTI-S$^\dag$~\cite{Monnier2021ICCV} & 89.86 & 53.38 & 79.86 & 32.20   \\
GNM$^\dag$~\cite{Jiang2020NeurIPS} & 65.67 & 63.38 & 53.38 & 44.30   \\
PPMP$^\dag$~\cite{Karazija2022NeurIPS} & \underline{95.94} & \underline{84.86} & \underline{92.61} & \textbf{77.67}   \\
Ours$^\dag$  & \textbf{96.95} & \textbf{86.38} & \textbf{95.32} & \underline{70.28} \\
\bottomrule    
\end{tabular}
\caption{\textbf{Segmentation Results on CLEVR and C{\footnotesize LEVR}T{\footnotesize EX} Datasets.} The lower part with $\dag$ shows the results with post-processing. The best result in each column is shown in \textbf{bold}, and the second best is \underline{underlined}. }
\label{tab:clevr-long}
\end{table}

%% file: figs/davis17.tex
\begin{figure*}[t]
    \centering
     \rotatebox[origin=l]{90}{
 \parbox{0.07\linewidth}{\centering RGB}}
 \includegraphics[width=0.155\linewidth]{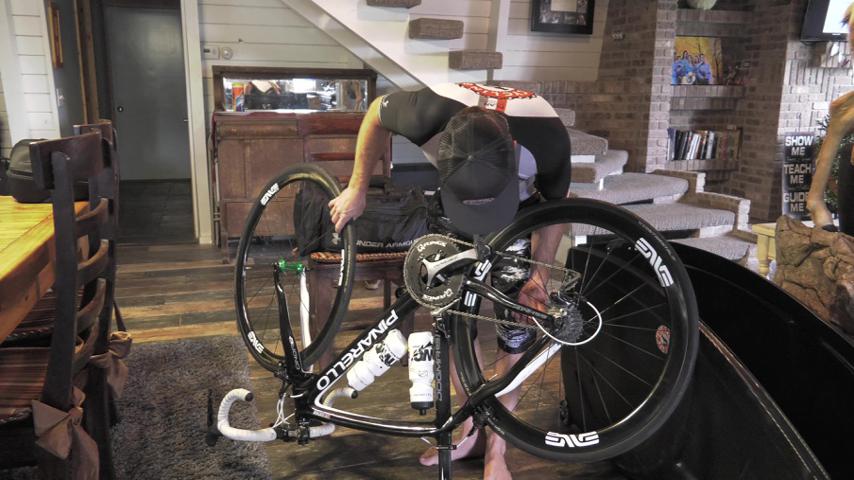}
    \includegraphics[width=0.155\linewidth]{gfx/noclusteringDavis17/india-00011.jpg}
    \includegraphics[width=0.155\linewidth]{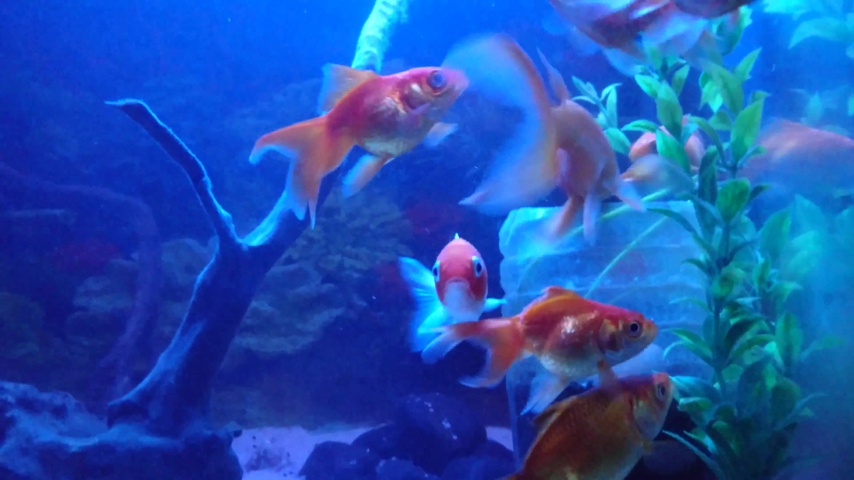}
    \includegraphics[width=0.155\linewidth]{gfx/noclusteringDavis17/judo-00028.jpg}
    \includegraphics[width=0.155\linewidth]{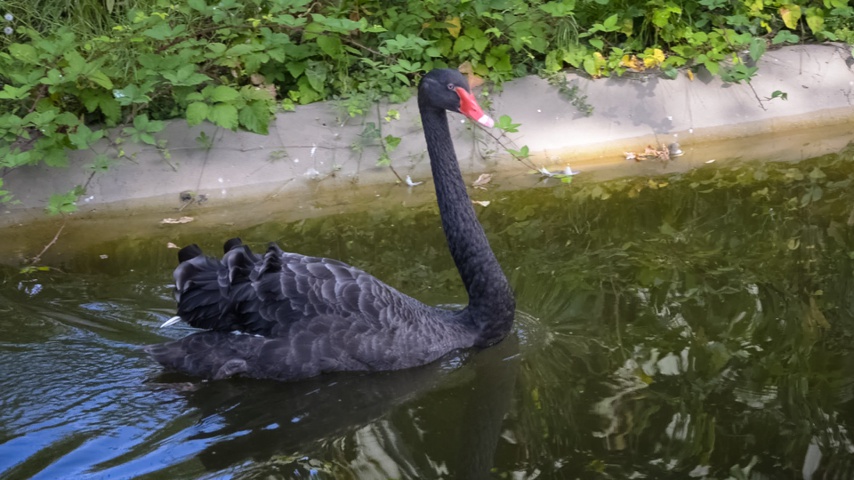}
    \includegraphics[width=0.155\linewidth]{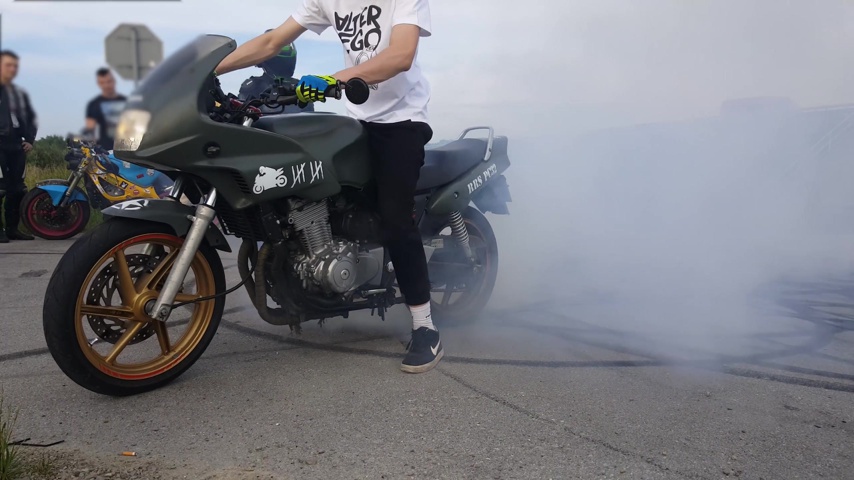}

    \rotatebox[origin=l]{90}{
 \parbox{0.07\linewidth}{\centering Flow}}
 \includegraphics[width=0.155\linewidth]{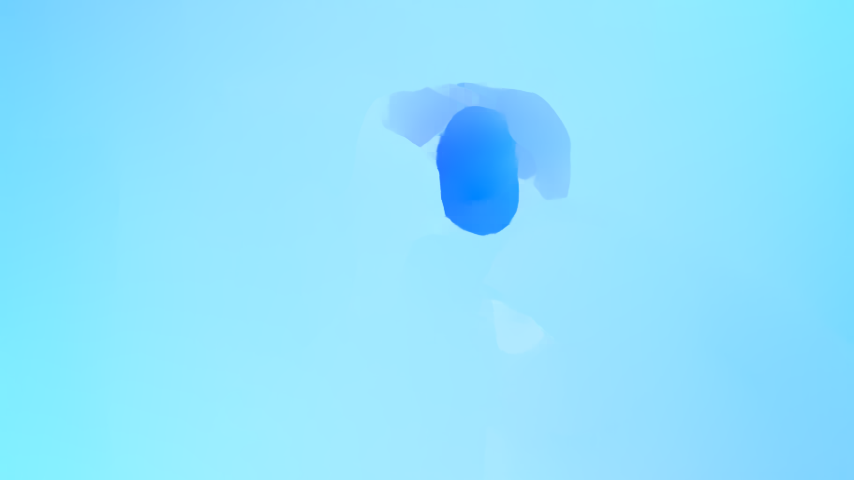}
    \includegraphics[width=0.155\linewidth]{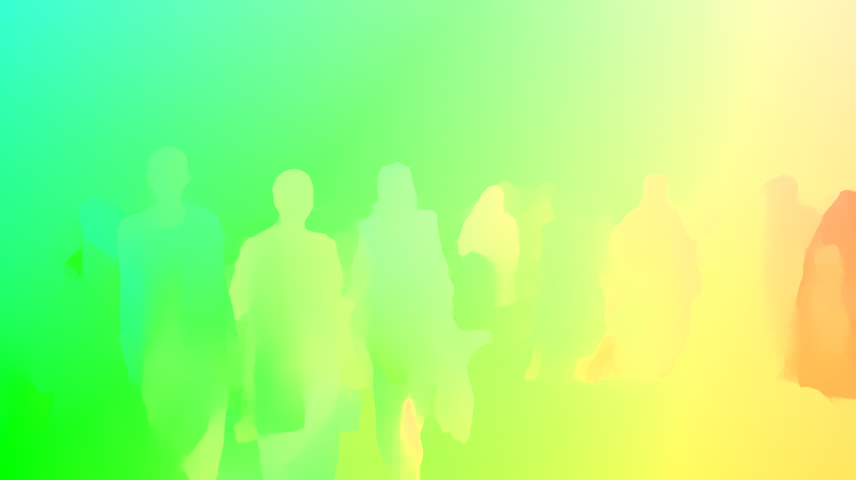}
    \includegraphics[width=0.155\linewidth]{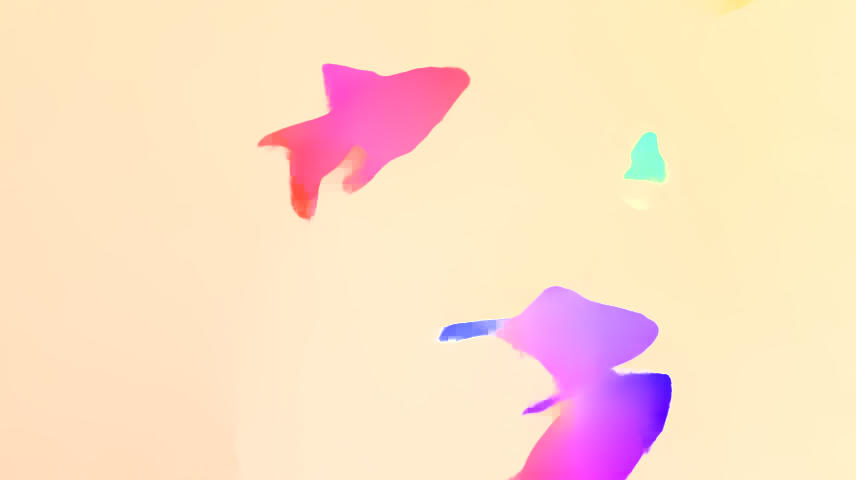}
    \includegraphics[width=0.155\linewidth]{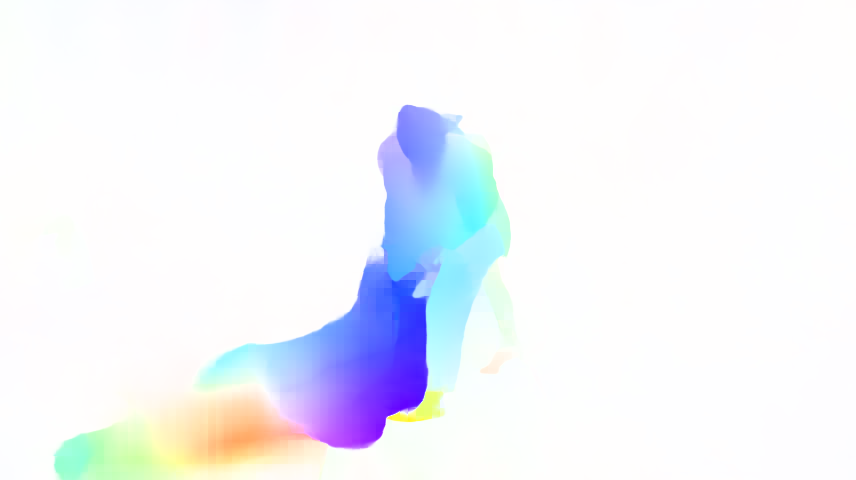}
    \includegraphics[width=0.155\linewidth]{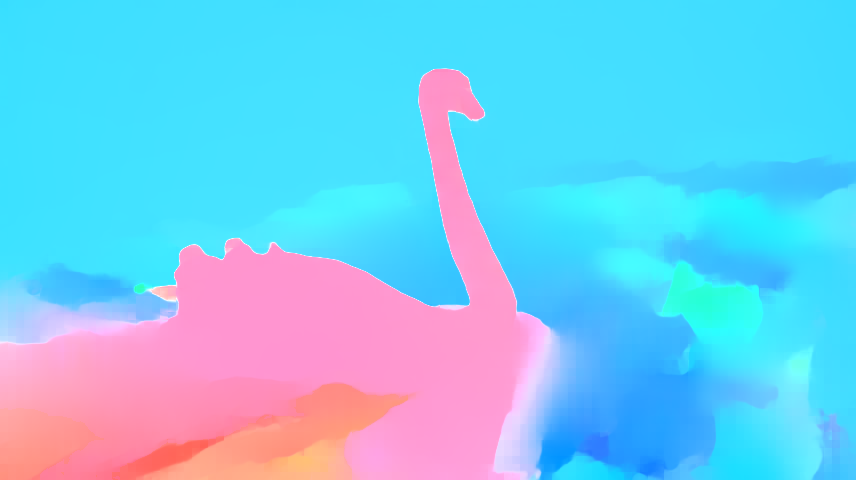}
    \includegraphics[width=0.155\linewidth]{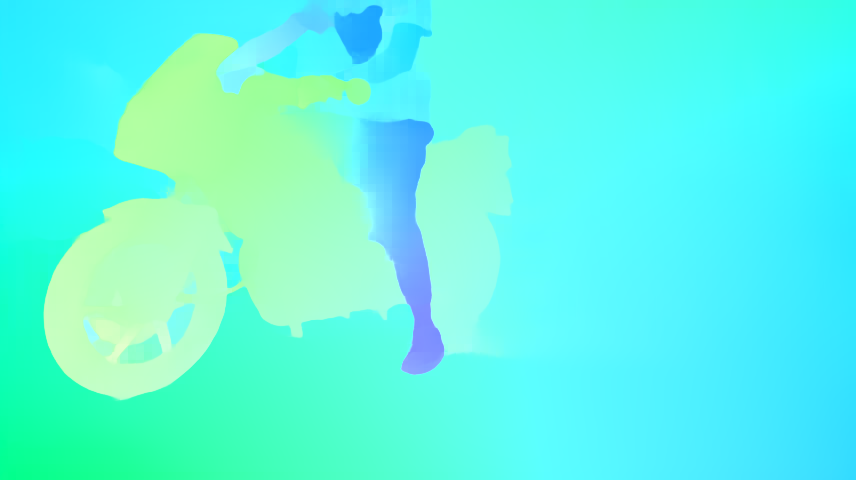}

    \rotatebox[origin=l]{90}{
 \parbox{0.07\linewidth}{\centering GT}}
 \includegraphics[width=0.155\linewidth]{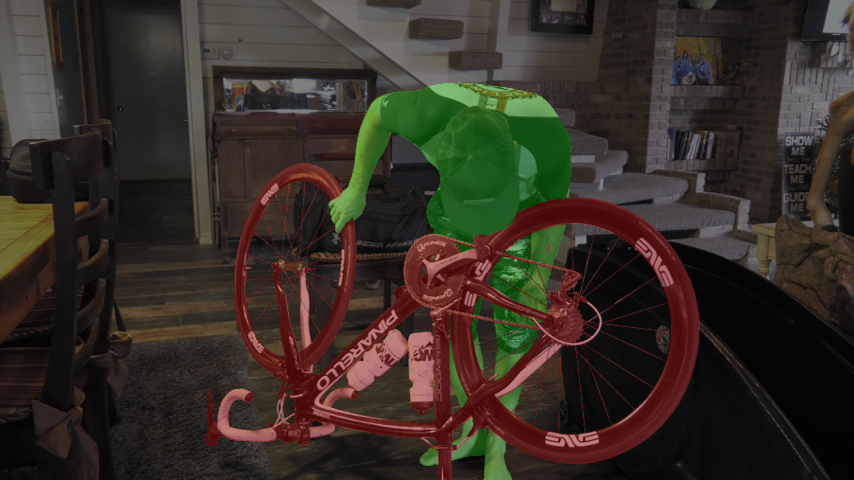}
    \includegraphics[width=0.155\linewidth]{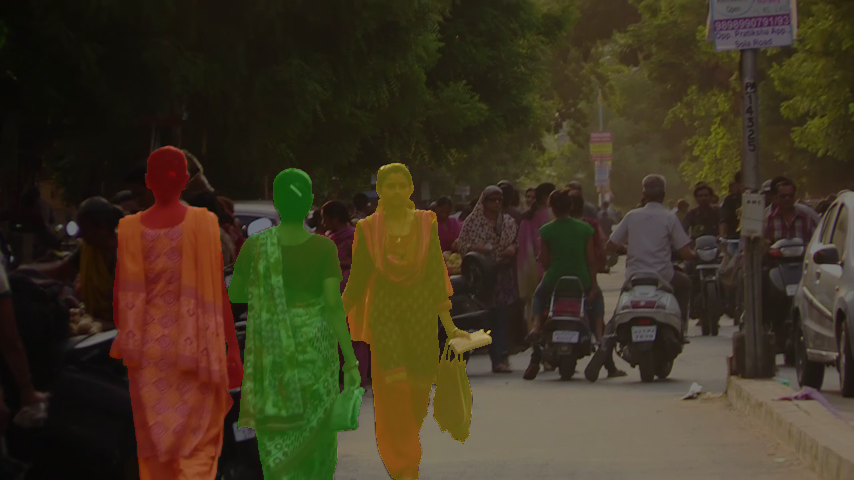}
    \includegraphics[width=0.155\linewidth]{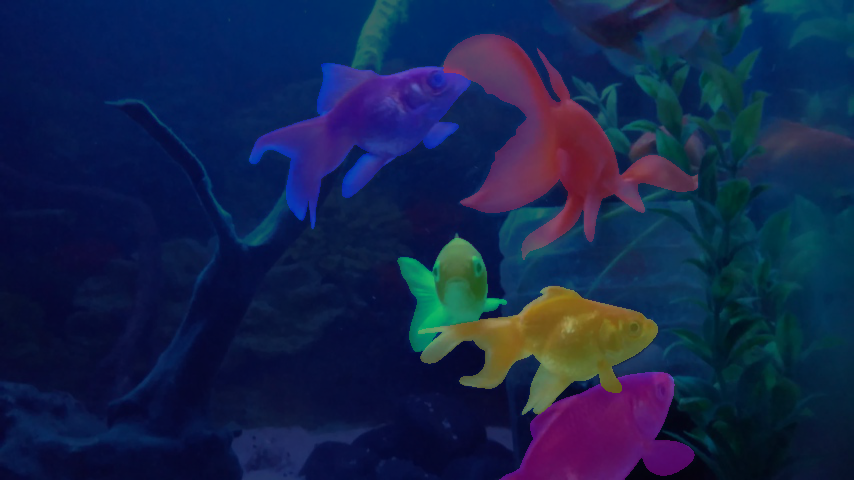}
    \includegraphics[width=0.155\linewidth]{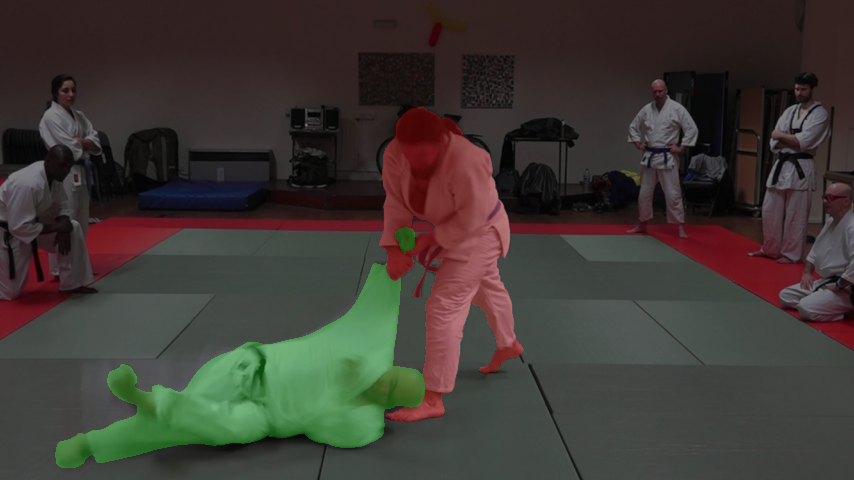}
    \includegraphics[width=0.155\linewidth]{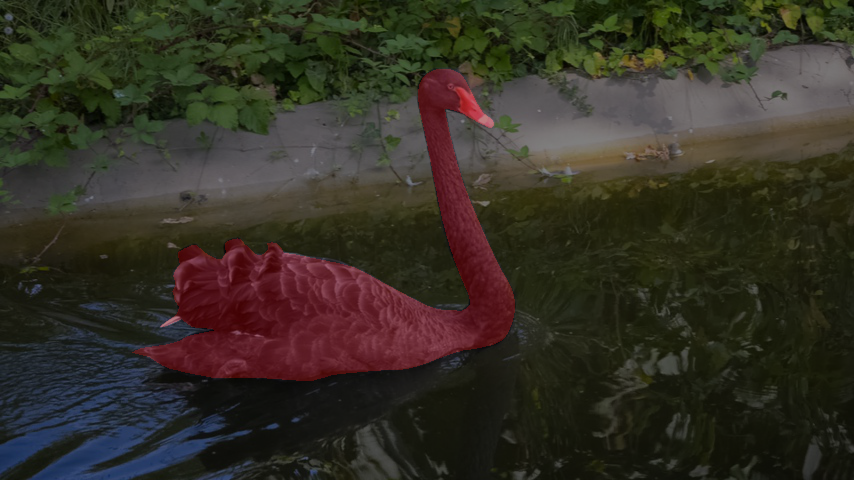}
    \includegraphics[width=0.155\linewidth]{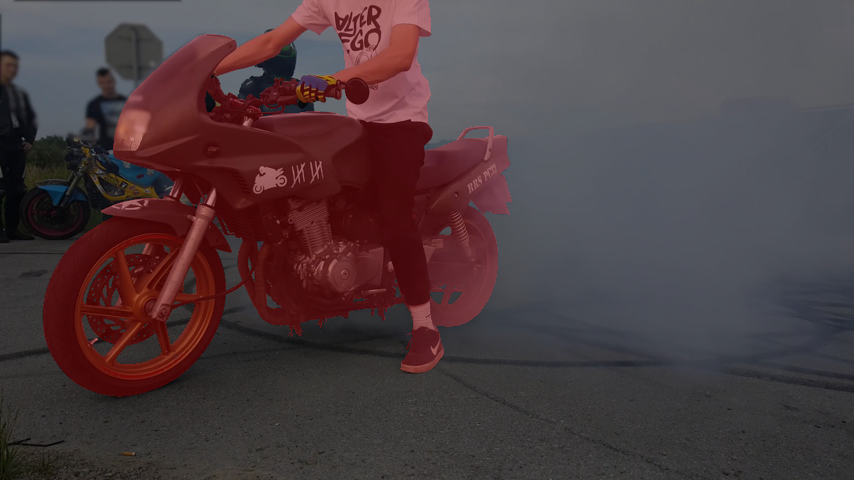}

    \rotatebox[origin=l]{90}{
 \parbox{0.07\linewidth}{\centering OCLR}}
 \includegraphics[width=0.155\linewidth]{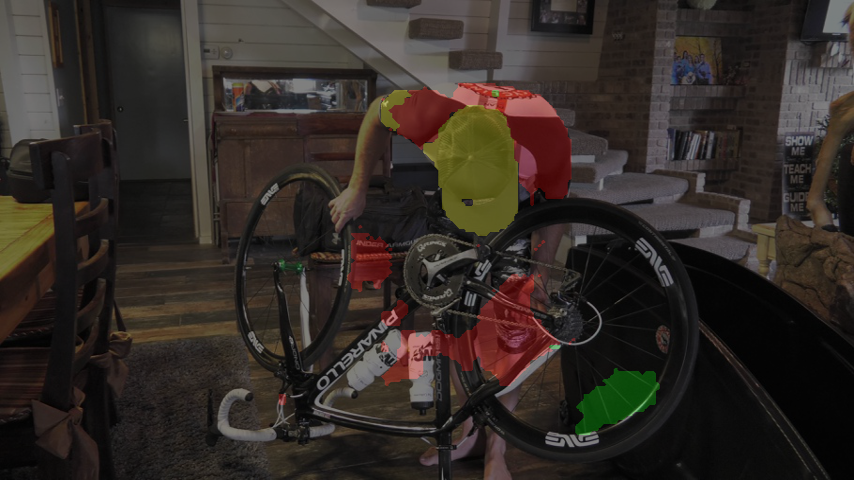}
    \includegraphics[width=0.155\linewidth]{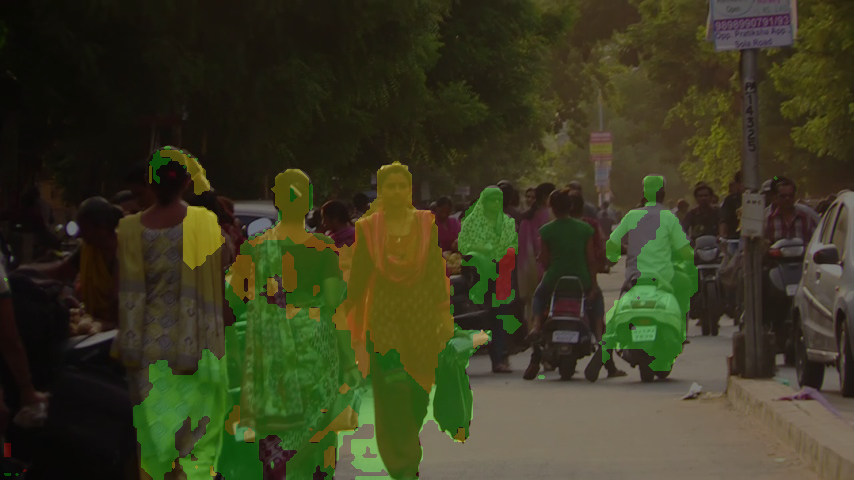}
    \includegraphics[width=0.155\linewidth]{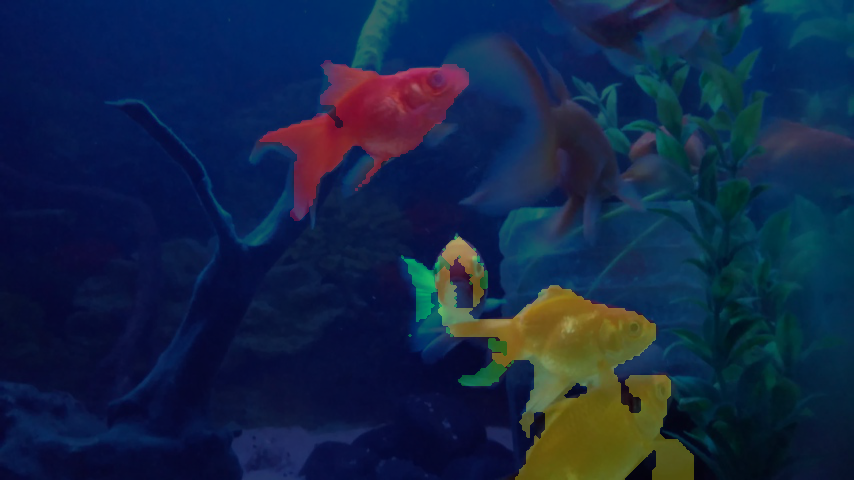}
    \includegraphics[width=0.155\linewidth]{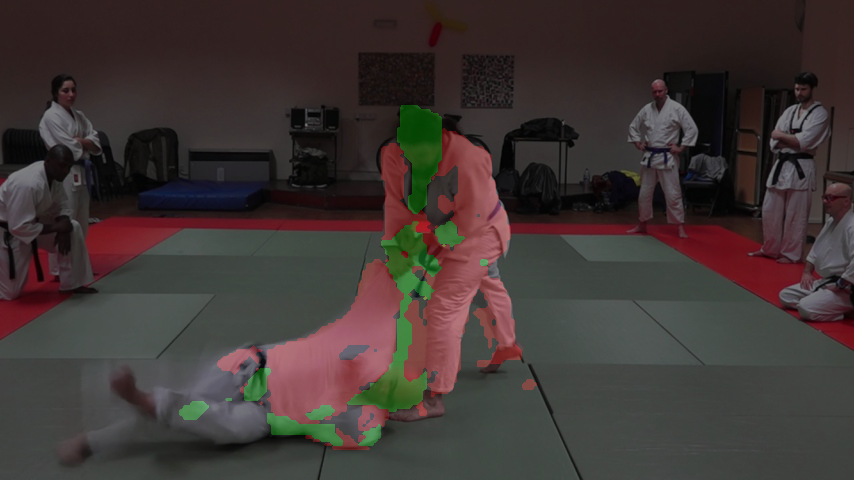}
    \includegraphics[width=0.155\linewidth]{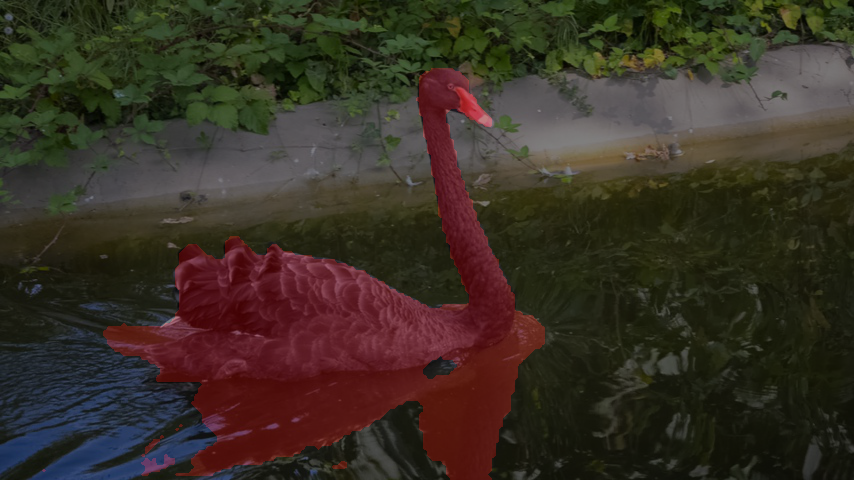}
    \includegraphics[width=0.155\linewidth]{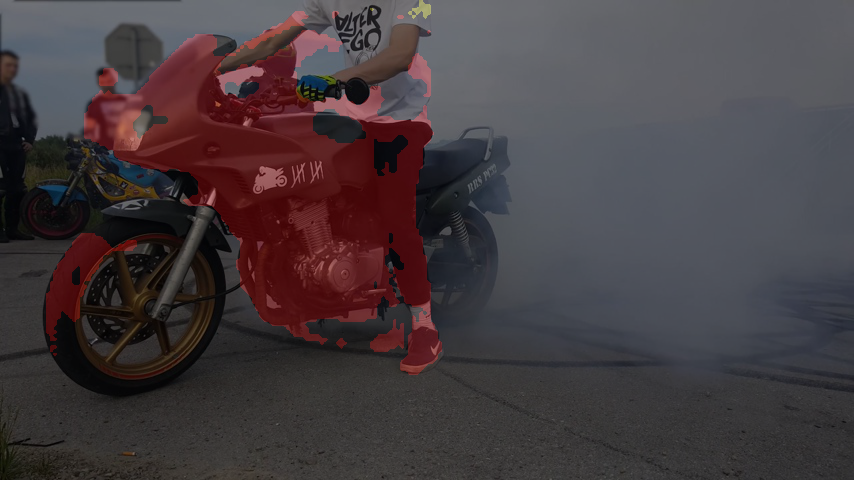}

    \rotatebox[origin=l]{90}{
 \parbox{0.07\linewidth}{\centering Ours}}
 \includegraphics[width=0.155\linewidth]{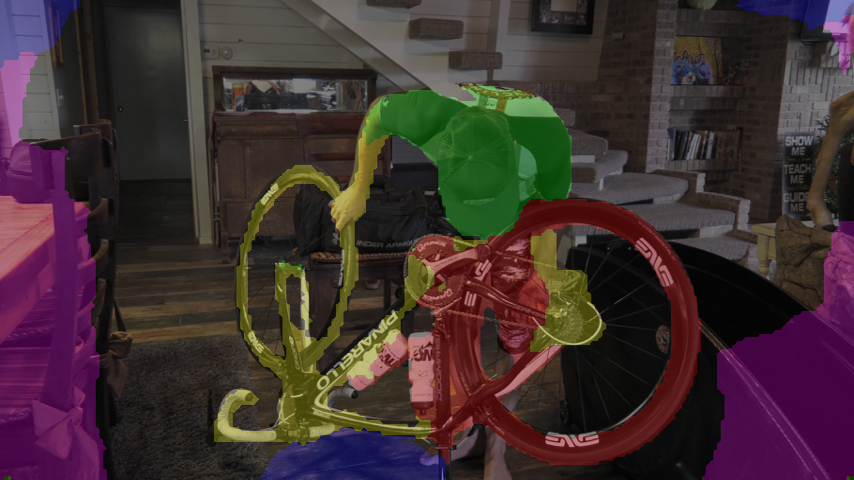}
    \includegraphics[width=0.155\linewidth]{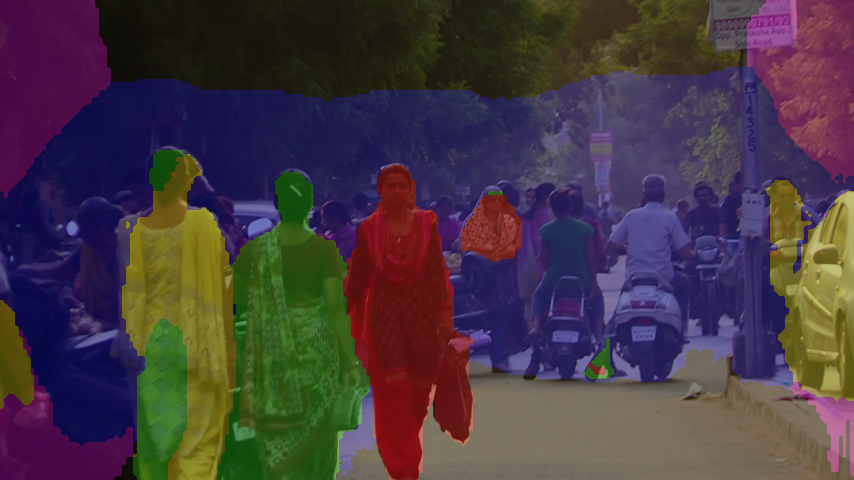}
    \includegraphics[width=0.155\linewidth]{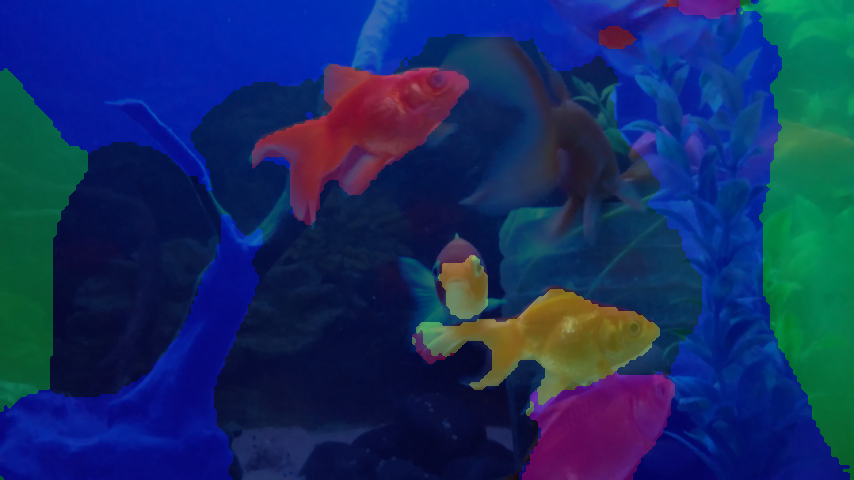}
    \includegraphics[width=0.155\linewidth]{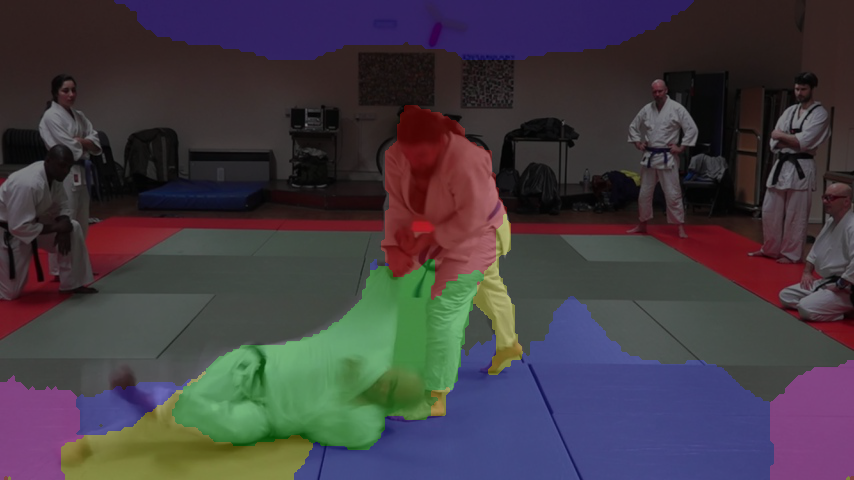}
    \includegraphics[width=0.155\linewidth]{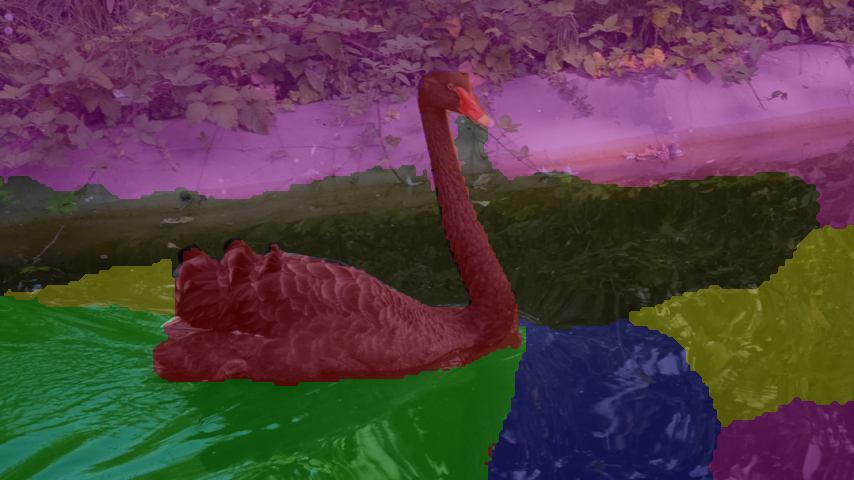}
    \includegraphics[width=0.155\linewidth]{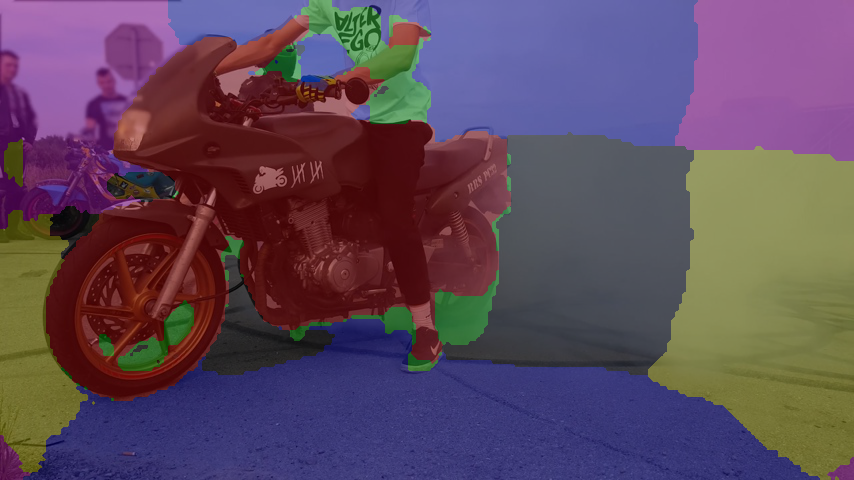}

    \rotatebox[origin=l]{90}{
 \parbox{0.07\linewidth}{\centering Ours-M}}
 \includegraphics[width=0.155\linewidth]{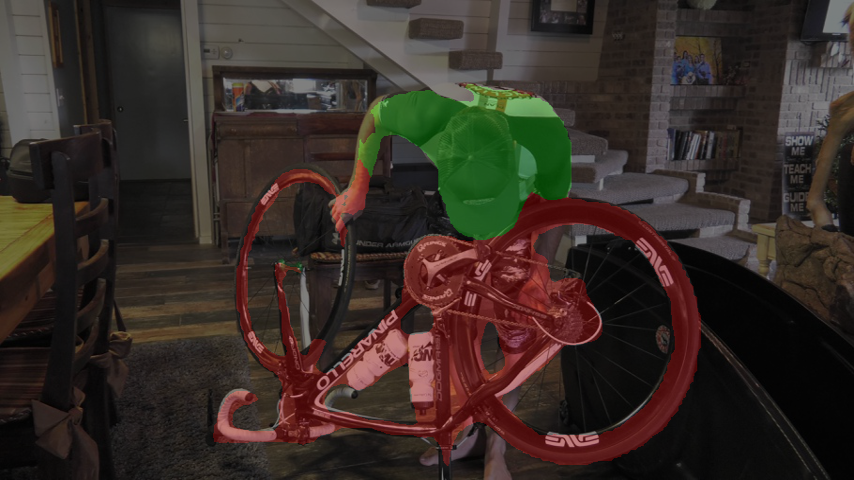}
    \includegraphics[width=0.155\linewidth]{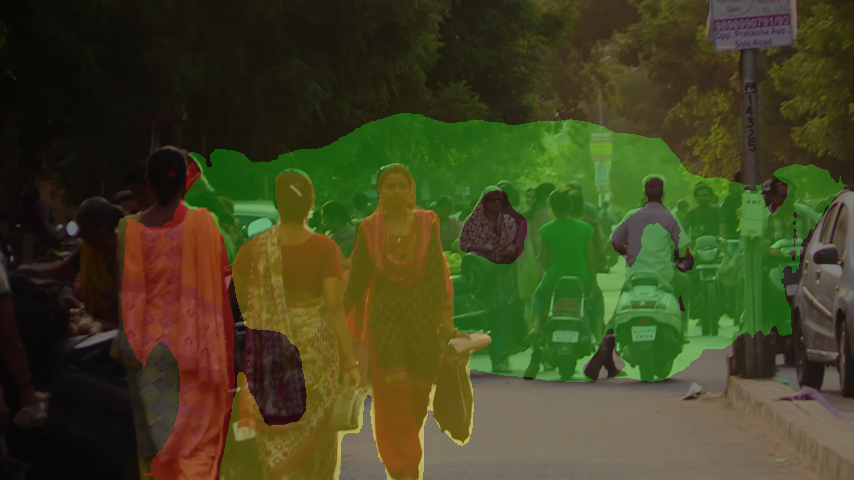}
    \includegraphics[width=0.155\linewidth]{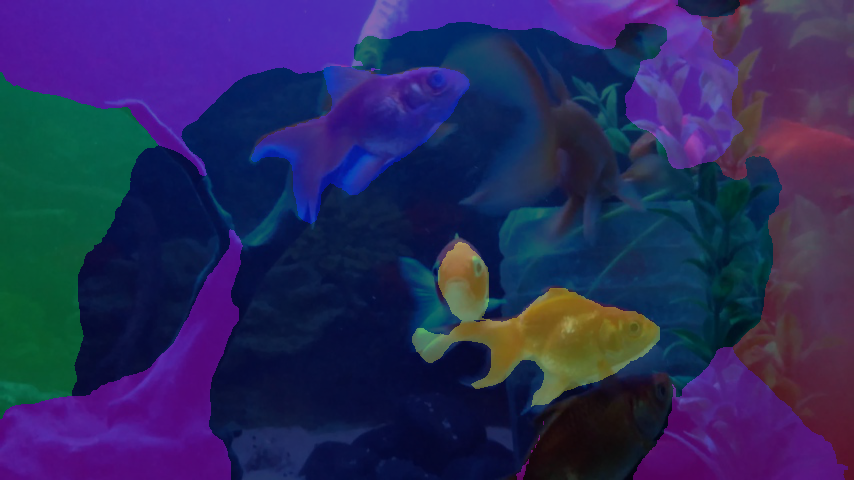}
    \includegraphics[width=0.155\linewidth]{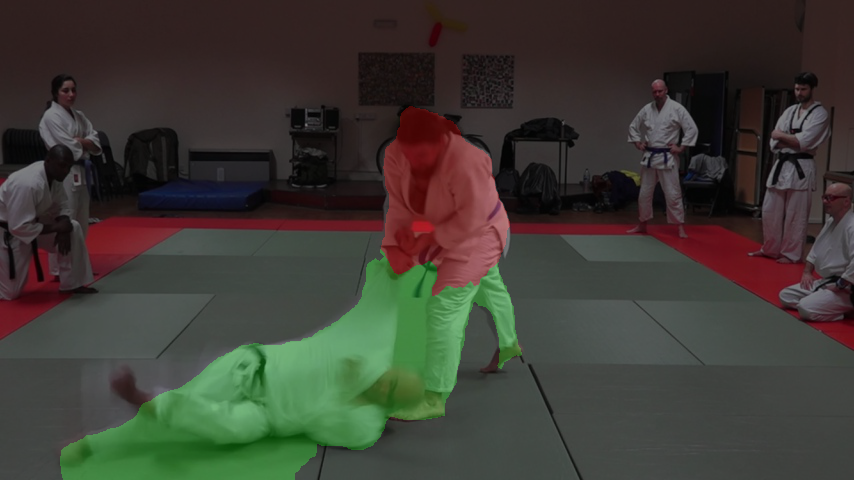}
    \includegraphics[width=0.155\linewidth]{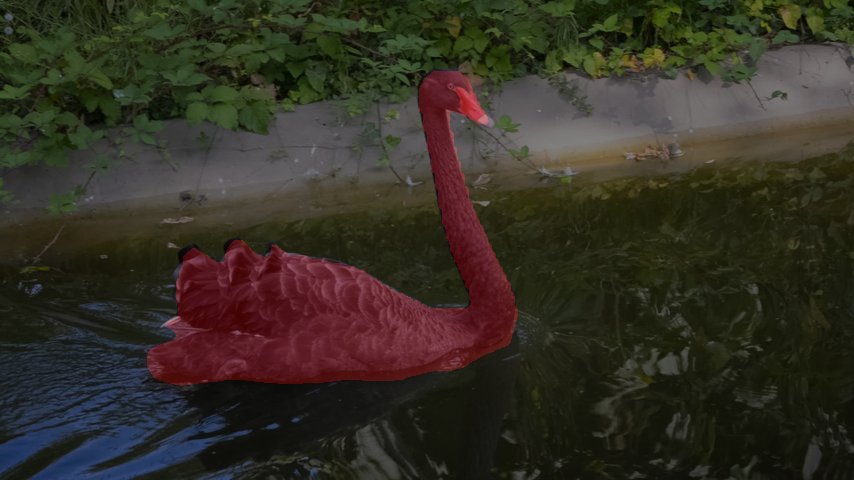}
    \includegraphics[width=0.155\linewidth]{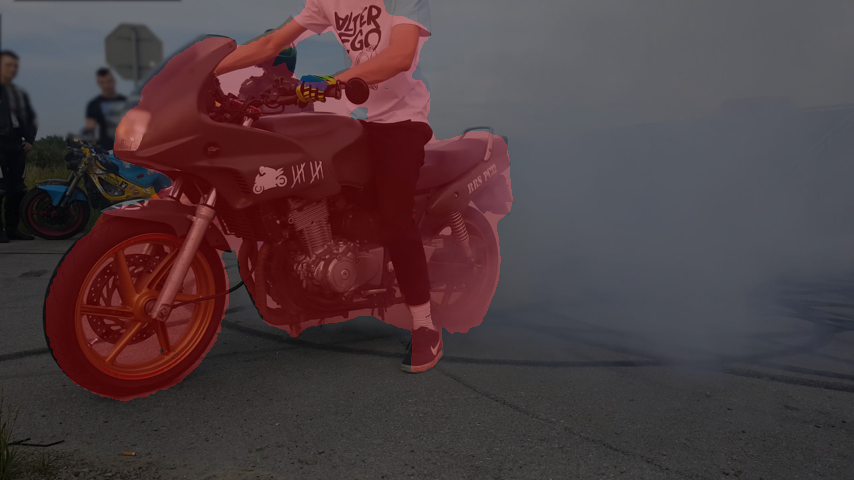}

    \caption{\textbf{Qualitative Comparison on DAVIS-2017}.  
    While OCLR~\cite{Xie2022NeurIPS} misses some objects completely and suffers from relying on only optical flow as input, our method can segment a wide variety of multiple objects in everyday scenes.}
    \label{fig:davis17}
    \vspace{-0.3cm}
\end{figure*}

%% file: tab/davis17.tex
\begin{table}[b]
\centering
\begin{tabular}{@{}lccc@{}}
\toprule
Model    & $\mathcal{J}$\& $\mathcal{F}$ $\uparrow$ & $\mathcal{J}$ $\uparrow$ & $\mathcal{F}$ $\uparrow$ \\ \midrule
Motion Grouping~\cite{Yang2021ICCV} & 35.8 & 38.4  & 33.2 \\
Motion Grouping (sup.) & 39.5 & 44.9  & 34.2 \\
Mask R-CNN (flow) & 50.3 & 50.4  & 50.2 \\
OCLR~\cite{Xie2022NeurIPS} & 55.1 & 54.5  & \underline{55.7} \\
Ours & \underline{55.3} & \underline{55.3}  & 55.3 \\
Ours-M & \textbf{59.2} & \textbf{59.3}  & \textbf{59.2} \\
\bottomrule
\end{tabular}
\caption{\textbf{Multi-Object Segmentation Results on DAVIS-2017.} We evaluate by using the motion labels from \cite{Xie2022NeurIPS}.}
\label{tab:davis}
\end{table}

%% file: tab/kitti_combined.tex
\begin{table*}[bt!]
\begin{subtable}[]{0.2\textwidth}
\centering
\footnotesize
\begin{tabular}{@{}lccc@{}}
\toprule
Model    & \textbf{FG-ARI} $\uparrow$ \\ \midrule
SA~\cite{Locatello2020NeurIPS} & 13.8 \\
MONet~\cite{Burgess2019ARXIV} & 14.9 \\
SCALOR~\cite{Jiang2020ICLR} & 21.1 \\
S-IODINE~\cite{Greff2019ICML} & 14.4 \\
MCG~\cite{Arbelaez2014CVPR}  & 40.9 \\
Ours & 42.3 \\
Bao \etal ~\cite{Bao2022CVPR} & 47.1 \\
PPMP~\cite{Karazija2022NeurIPS} & 50.8 \\
\bottomrule
\end{tabular}
\caption{Segmentation}
\label{tab:seg-kitti}
\end{subtable}
\hspace{0.3cm}
\begin{subtable}[]{0.8\textwidth}
\centering
\footnotesize
\begin{tabular}{@{}lccccccc@{}}
\toprule
 & Abs Rel $\downarrow$ & Sq Rel $\downarrow$ & RSME $\downarrow$ & RMSE log $\downarrow$ & $\delta < 1.25$ $\uparrow$ & $ \delta < 1.25^2$ $\uparrow$ & $ \delta < 1.25^3$ $\uparrow$ \\ \midrule
Zhou \etal ~\cite{Zhou2017CVPR} & 0.176 & 1.532& 6.129 & 0.244 & 0.758 & 0.921 & 0.971 \\
Mahjourian \etal ~\cite{Mahjourian2018CVPR} & 0.134 & 0.983& 5.501 & 0.203 & 0.827 & 0.944 & 0.981 \\
GeoNet~\cite{Yin2018CVPR} & 0.132 & 0.994& 5.240 & 0.193 & 0.833 & 0.953 & 0.985 \\
DDVO~\cite{Wang2018CVPR} & 0.126 & 0.866& 4.932 & 0.185 & 0.851 & 0.958 & 0.986 \\
Ranjan \etal ~\cite{Ranjan2019CVPR} & 0.123 & 0.881& 4.834 & 0.181 & 0.860 & 0.959 & 0.985 \\
EPC++~\cite{Luo2019PAMI} & 0.120 & 0.789& 4.755 & 0.177 & 0.856 & 0.961 & 0.987 \\
Ours & 0.107 & 1.539& 4.027 & 0.149 & 0.911 & 0.971 & 0.989 \\
Monodepth2~\cite{Godard2019ICCV} & 0.090 & 0.545& 3.942 & 0.137 & 0.914 & 0.983 & 0.998 \\
PackNet-SfM~\cite{Guizilini2020CVPR} & 0.078 & 0.420& 3.485 & 0.121 & 0.934 & 0.986 & 0.996 \\
\bottomrule
\end{tabular}
\caption{Depth}
\label{tab:depth-kitti}
\end{subtable}
\vspace{-0.3cm}
\caption{\textbf{Results on KITTI.} We evaluate the segmentation on the instance segmentation benchmark, and the depth on the KITTI Eigen split~\cite{Eigen2014NeurIPS} with improved ground truth~\cite{Uhrig2017THREEDV}.}
\label{tab:kitti}
\end{table*}

%% file: tab/ablation_movi.tex
\begin{table*}[ht!]
\centering
\footnotesize
\begin{tabular}{@{}lcccccccc@{}}
\toprule
  & \multicolumn{2}{c}{MOVi-A}    & \multicolumn{2}{c}{MOVi-C}    & \multicolumn{2}{c}{MOVi-D}    & \multicolumn{2}{c}{MOVi-E}      \\ \cmidrule(lr){2-3} \cmidrule(lr){4-5} \cmidrule(lr){6-7}\cmidrule(l){8-9}
  & \textbf{FG-ARI}$\uparrow$  & \textbf{mIoU}$\uparrow$  & \textbf{FG-ARI}$\uparrow$  & \textbf{mIoU}$\uparrow$  & \textbf{FG-ARI}$\uparrow$  & \textbf{mIoU}$\uparrow$  & \textbf{FG-ARI}$\uparrow$  & \textbf{mIoU}$\uparrow$  \\ \midrule 
Only-T     & 31.13 & 17.72 & 69.69 &  25.84 & 68.68 & 22.74 & 64.69 & 28.26\\
Only-R     & \textbf{90.43} & \textbf{75.63} & 68.51 &  53.00 & 70.57 & 56.97 & 73.03 & 40.34\\
Full     & 56.09 & 36.48 & \textbf{73.80} & \textbf{54.48} & \textbf{76.41} & \textbf{58.82} & \textbf{78.33} & \textbf{47.38} \\
\midrule
Only-T$^{\dag}$ & 47.96 & 29.92 & 69.94 & 26.31 & 68.66 & 22.76 & 69.94 & 30.71 \\
Only-R$^\dag$  & \textbf{92.09} & \textbf{84.60} & 69.23 &  55.54 & 71.20 & 58.41 & 76.20 & 42.96\\
Full$^\dag$  & 70.15 & 46.26 & \textbf{74.64} & \textbf{59.24} & \textbf{77.15} & \textbf{59.68} & \textbf{80.83} & \textbf{50.48} \\
\bottomrule
\end{tabular}
\caption{\textbf{Ablation Study.} We perform an ablation study by using only the translation (Only-T) or rotation (Only-R) component and compare it to our model with both (Full). See text for details.}
\vspace{-0.3cm}
\label{tab:movi_ablation}
\end{table*}

%% file: sec/conclusion.tex
\section{Conclusion and Future Work}
We presented a motion-supervised approach for multi-object segmentation that can work with a single RGB image at test time, therefore still applicable to image datasets. Our method is the first to consider geometry to remove ambiguity for multi-object segmentation from a single image without using any labels for segmentation. Modeling geometry significantly advances the state-of-the-art on commonly used synthetic datasets. We also evaluated our method on real-world datasets. 
Our method is the first image-based multi-object segmentation method to report state-of-the-art results on DAVIS-2017 without using motion at test time. We also report comparable results for depth prediction on KITTI and MOVi datasets where depth can be evaluated.

Predicting objects that can potentially move independently from a single image requires observing examples of various objects moving in the training set. Moreover, static objects send a mixed signal to the model. The coherent changes in the flow can be captured with the help of geometry as shown in our work. The remaining uncertainty can be addressed with a probabilistic formulation as done in previous state-of-the-art~\cite{Karazija2022NeurIPS}. 
Another problem is scenes without enough information to predict depth as we observed on textureless MOVi-A. However, the lack of information to this extent rarely happens on real-world data.

%% file: sec/supp/summary.tex
In this supplementary document, we first provide the derivations of the basis for the space of possible optical flows in \secref{sec:supp-basis}. Then in \secref{sec:supp-project}, we provide the details of the projection of the input flow into the space spanned by the bases. In \secref{sec:supp-qual}, we show additional qualitative results and in \secref{sec:supp-movi-depth}, we provide an evaluation of our depth estimations for the foreground objects on MOVi datasets. Finally, in \secref{sec:supp-intrinsics}, we show depth evaluation results for our model assuming known camera intrinsics.

%% file: sec/supp/basis.tex
\section{Derivation of Basis}
\label{sec:supp-basis}
Assume that the world coordinate system coincides with the camera coordinate system and let $\bX = (\bx, \by, \bz)$ denote the coordinates of a 3D point in the world. Assume that the scene is static and the camera is moving rigidly with angular velocity $\omega \in \nR^3$ and linear velocity $v \in \nR^3$, corresponding to the rotational and translational part of the motion. Then, following \cite{Heeger1992IJCV, Ma2003Invitation}, $\bX'$, the instantaneous velocity of the point $\bX$, can be calculated as follows: 

\begin{equation}
\label{eq:3Dvelocity}
    \bX' = 
    \begin{bmatrix}
        \bx' \\ \by' \\ \bz'
    \end{bmatrix} = -(\omega \times \bX + v) =
    \begin{bmatrix}
    \omega_3 \by - \omega_2 \bz -  v_1\\
    \omega_1 \bz - \omega_3 \bx -  v_2\\
    \omega_2 \bx - \omega_1 \by  - v_3
    \end{bmatrix}
\end{equation}

Let $f_x, f_y$ be the focal lengths and $(c_x, c_y)$ denote the principal point of the camera.  The pixel $\bp = [u, v]^T$ corresponding to the 3D point $\bX$ can be calculated as:
\begin{equation}
\label{eq:X2p}
    \bp = \begin{bmatrix}
        u \\
        v
    \end{bmatrix} = 
    \begin{bmatrix}
        {\bx f_x}/{\bz} + c_x \\
        {\by f_y}/{\bz} + c_y
    \end{bmatrix}
\end{equation}
Therefore, we can write:
\begin{eqnarray}
\label{eq:simplify}
    \dfrac{\bx}{\bz} = \dfrac{(u-c_x)}{f_x} = f_x^{-1}\bar{u} \nonumber \\
    \dfrac{\by}{\bz} = \dfrac{(v-c_y)}{f_y} = f_y^{-1}\bar{v}
\end{eqnarray}
where we have defined $\bar{u} = u-c_x$ and $\bar{v} = v-c_y$.
The instantaneous flow of a pixel $\bp$ can be computed by taking derivatives of \eqnref{eq:X2p} with respect to time as follows:
\begin{equation}
\label{eq:2Dvelocity}
 \bp' = \begin{bmatrix}
        u' \\
        v'
    \end{bmatrix} = 
     \dfrac{1}{\bz^2}\begin{bmatrix}
        {f_x(\bz \bx' - \bx\bz')} \\
        {f_y(\bz \by'- \by\bz')}
    \end{bmatrix}
\end{equation}
By substituting the values from \eqnref{eq:3Dvelocity} into \eqnref{eq:2Dvelocity} we can write:
\begin{align}
\label{eq:long}
\begin{bmatrix}
        u' \\
        v'
    \end{bmatrix} &= 
    \dfrac{1}{\bz^2}\begin{bmatrix}
        f_x \left(\bz \left(\omega_3 \by - \omega_2 \bz -  v_1 \right) - \bx \left(\omega_2 \bx - \omega_1 \by  - v_3 \right) \right) \\
        f_y \left( \bz\left(\omega_1 \bz - \omega_3 \bx -  v_2 \right) - \by \left(\omega_2 \bx - \omega_1 \by  - v_3 \right) \right)
    \end{bmatrix} \nonumber \\ 
    &= 
    \dfrac{1}{\bz^2}\begin{bmatrix}
         f_x(-\bz v_1 + \bx v_3  + \bx\by \omega_1 - (\bx^2 + \bz^2) \omega_2 + \by\bz \omega_3)
         \\
         f_y(-\bz v_2 + \by v_3 + (\by^2 + \bz^2)\omega_1 -\bx\by \omega_2 - \bx\bz \omega_3)
    \end{bmatrix}
\end{align} 
By plugging the values from \eqnref{eq:simplify}, and using disparity $d=1/\bz$, we can re-write \eqnref{eq:long} as:
\begin{align}
    \begin{bmatrix}
        u' \\
        v'
    \end{bmatrix}
    = \begin{bmatrix}
       -f_x~d      &  0  \\
       0          &  -f_y~d  \\
       \bar{u}~d &  \bar{v}~d  \\
       {f_y}^{-1}~\bar{u}~\bar{v}   & f_y + {f_y}^{-1}~\bar{v}^2  \\
       -(f_x + {f_x}^{-1}~\bar{u}^2) & -{f_x}^{-1}~\bar{u}~\bar{v}    \\
       f_x~{f_y}^{-1}~\bar{v} &   - f_y~{f_x}^{-1}~\bar{u} 
    \end{bmatrix}^T
    \begin{bmatrix}
        v_1 \\
        v_2 \\
        v_3 \\
        \omega_1 \\
        \omega_2 \\
        \omega_3
    \end{bmatrix}
\end{align}

Therefore, we can define a basis for the space of possible instantaneous optical flows for a given frame as
\begin{equation}
\label{eq:six-basis}
\mathcal{B}_0 = \{\bb_{\bT x}, \bb_{\bT y}, \bb_{\bT z}, \bb_{\bR x}, \bb_{\bR y}, \bb_{\bR z}\}
\end{equation}
where we define:
\begin{eqnarray}
        &\bb_{\bT x} = \begin{bmatrix}
              f_x~d \\
             0
          \end{bmatrix}, 
        &\bb_{\bR x} =  \begin{bmatrix}
              {f_y}^{-1}~\bar{u}~\bar{v} \\
             f_y + {f_y}^{-1}~\bar{v}^2
          \end{bmatrix} \nonumber \\
        \nonumber \\ %
        &\bb_{\bT y} = \begin{bmatrix}
              0 \\
             f_y~d
          \end{bmatrix}, 
        &\bb_{\bR y} = \begin{bmatrix}
            f_x + {f_x}^{-1}~\bar{u}^2 \\
            {f_x}^{-1}~\bar{u}~\bar{v}
        \end{bmatrix} \nonumber \\
        \nonumber \\ %
        &\bb_{\bT z} = \begin{bmatrix}
             -\bar{u}~d \\
             -\bar{v}~d
          \end{bmatrix},
        &\bb_{\bR z} = \begin{bmatrix}
             f_x~{f_y}^{-1}~\bar{v} \\
             -f_y~{f_x}^{-1}~\bar{u}
          \end{bmatrix}  
\end{eqnarray}

Our goal is to have basis vectors that do not depend on the values of focal lengths. Since  basis vectors can be scaled arbitrarily, we can scale $\bb_{\bT x}$ and $\bb_{\bT y}$ by $1/f_x$ and $1/f_y$, respectively, to make them independent of $f_x$ and $f_y$. By assuming $f_x = f_y$, $\bb_{\bR z}$ becomes $[ \bar{v}, -\bar{u}]^T$ which is also free of focal lengths. We can write $\bb_{\bR x}$ and $\bb_{\bR y}$ as:
\begin{align}
    \bb_{\bR x} & = f_y \begin{bmatrix}
        0 \\ 1
    \end{bmatrix} + {f_y}^{-1} \begin{bmatrix}
        \bar{u}~\bar{v} \\ \bar{v}^2
    \end{bmatrix} \nonumber \\
    \bb_{\bR y} & = f_x \begin{bmatrix}
        1 \\ 0
    \end{bmatrix} + {f_x}^{-1} \begin{bmatrix}
        \bar{u}^2 \\ \bar{u}~\bar{v}
    \end{bmatrix}
\end{align}
Therefore, if we define: 
\begin{eqnarray}
    \bb_{\bR^1 x} =  \begin{bmatrix}
        0 \\ 1 
    \end{bmatrix} & \bb_{\bR^2 x} = \begin{bmatrix}
        \bar{u}~\bar{v} \\ \bar{v}^2
    \end{bmatrix} \nonumber \\
    \bb_{\bR^1 y} =  \begin{bmatrix}
        1 \\ 0 
    \end{bmatrix} & \bb_{\bR^2 y} = \begin{bmatrix}
        \bar{u}^2 \\ \bar{u}~\bar{v}
    \end{bmatrix}
\end{eqnarray}
we can replace $\bb_{\bR x}$ with the pair $\bb_{\bR^1 x}$ and $\bb_{\bR^2 x}$. Similarly, we replace $\bb_{\bR y}$ with the pair $\bb_{\bR^1 y}$ and $\bb_{\bR^2 y}$~\cite{Bowen2022THREEDV}.
 Therefore, we can use the set of eight basis vectors
 \begin{equation}
 \label{eq:eight-basis}
 \mathcal{B}_0=\{\bb_{\bT x}, \bb_{\bT y}, \bb_{\bT z}, \bb_{\bR^1 x}, \bb_{\bR^2 x}, \bb_{\bR^1 y}, \bb_{\bR^2 y}, \bb_{\bR z} \}
 \end{equation}
 as a basis for the space of possible flows. Note that the space covered by this basis is actually slightly bigger because we cannot enforce the $f_x = f_y$ constraint in the decomposition of the rotational flows~\cite{Bowen2022THREEDV}.

 We normalize $\bb_{\bT x}, \bb_{\bT y},$ and $\bb_{\bT z}$ so that each vector has norm $2$ before multiplication by $d$, and normalize $\bb_{\bR^1 x}, \bb_{\bR^2 x}, \bb_{\bR^1 y}, \bb_{\bR^2 y},$ and $\bb_{\bR^1 z}$ to have norm $1$.

%% file: sec/supp/projection.tex
\section{Projection of Flow}
\label{sec:supp-project}
We project input flow $\mathbf{F}$ into $\text{span}(\{\mathcal{B}_1 \cup \mathcal{B}_2 \cup \ldots \cup \mathcal{B}_K\})$ where each $\mathcal{B}_i$ is a set of 8 vectors defined as:
\begin{equation}
\mathcal{B}_i = \{ \bm_i \bb \mid \bb \in \mathcal{B}_0 \}.
\end{equation}
Consider an aribtrary ordering on the elements of $\mathcal{B}_i$ and define ${\bv_i}^j$ as the $j$'th element in $\mathcal{B}_i$, reshaped into a $2HW$ vector.
We define the matrix $\bS_i \in \nR^{2HW \times 8}$ as:
\begin{equation}
    \bS_i = 
    \begin{bmatrix} \bv_i^1 \mid \bv_i^2 \mid \ldots \mid \bv_i^8
    \end{bmatrix}
\end{equation}
Then, we define $\bS \in \nR^{2HW \times 8K}$ as follows:
\begin{equation}
    \bS = \begin{bmatrix}
        \bS_1 \mid \bS_2 \mid \ldots \mid \bS_K
    \end{bmatrix}
\end{equation}
We calculate the singular value decomposition of $\bS$:
\begin{equation}
    \bS = \bU \mathbf{\Sigma} \bV^T
\end{equation}
The columns of $\bU$ corresponding to non-zero singular values of $\bS$ span the column space of $\bS$, \ie $\text{span}(\{\mathcal{B}_1 \cup \mathcal{B}_2 \cup \ldots \cup \mathcal{B}_K\})$. Since the columns of $\bU$ form an orthonormal set, we can project $\mathbf{F}$ into the column space of $\bS$ as follows:
\begin{equation}
    \mathbf{\hat{F}} = \bU' \bU'^T \mathbf{F}
\end{equation}
where $\bU'$ is the matrix whose columns are the columns of $\bU$ corresponding to non-zero singular values of $\bS$. In practice, we select columns of $\bU$ that correspond to singular values larger than $10^{-5}$.

%% file: sec/supp/qual.tex
\section{Additional Qualitative Results}
\label{sec:supp-qual}
Qualitative results for CLEVR and C{\footnotesize LEVR}T{\footnotesize EX} datasets are provided in \figref{fig:clevr}. We show additional visualizations, including the post-processing results for MOVi datasets in \figref{fig:movi-pp}. We can see that PPMP~\cite{Karazija2022NeurIPS} suffers from over-segmentation, with or without post-processing, especially in the MOVi datasets, whereas our method achieves much better results, as reflected in the quantitative performance. Our results for the KITTI dataset are visualized in \figref{fig:kitti}. It can be seen that we can segment objects such as cars and pedestrians successfully. We also visualize the depth estimations of our model. 
\input{figs/clevr}

\input{figs/movi_with_pp}
\input{figs/kitti}

%% file: figs/clevr.tex
\begin{figure*}[t]
    \centering
 
 \rotatebox[origin=l]{90}{
 \parbox{0.09\linewidth}{ Our depth}
 \parbox{0.09\linewidth}{\centering Ours$^{\dag}$}
 \parbox{0.09\linewidth}{\centering Ours}
 \parbox{0.09\linewidth}{\centering PPMP$^{\dag}$}
 \parbox{0.09\linewidth}{\centering PPMP}
 \parbox{0.09\linewidth}{\centering GT Seg}
 \parbox{0.09\linewidth}{\centering Flow}
 \parbox{0.09\linewidth}{\centering RGB}}
 \hspace{0.001\linewidth} \includegraphics[width=0.093\linewidth]{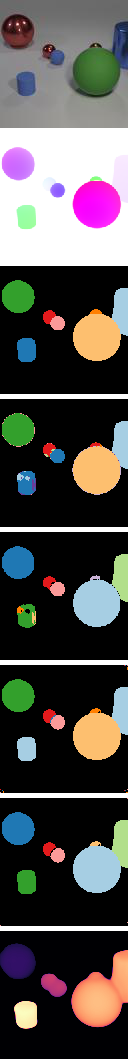}
    \includegraphics[width=0.093\linewidth]{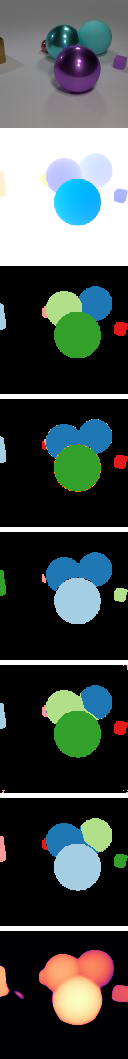}
    \includegraphics[width=0.093\linewidth]{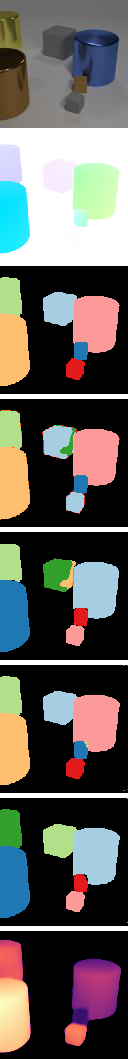}
    \includegraphics[width=0.093\linewidth]{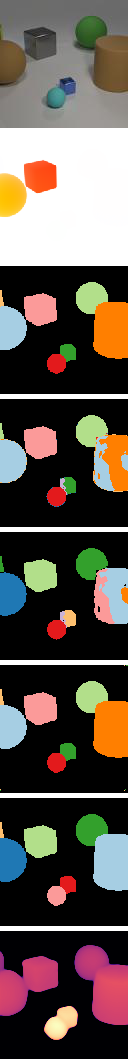}
    \hspace{0.002\linewidth}
    \includegraphics[width=0.093\linewidth]{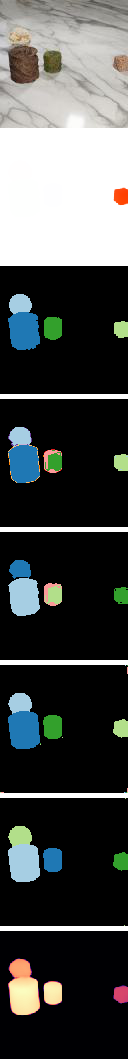}
    \includegraphics[width=0.093\linewidth]{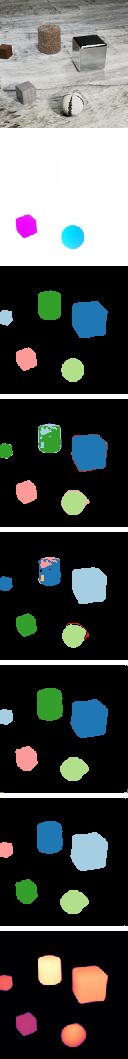}
    \includegraphics[width=0.093\linewidth]{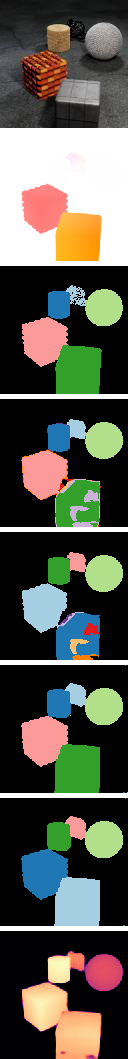}
    \includegraphics[width=0.093\linewidth]{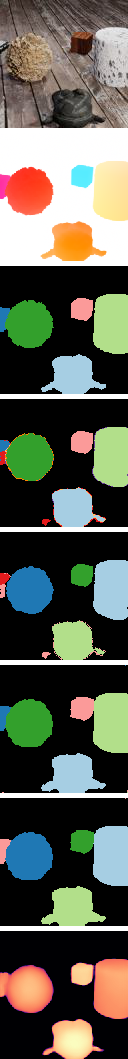}
    \caption{\textbf{Visualization of Depth and Segmentation Results on CLEVR and C{\footnotesize LEVR}T{\footnotesize EX} datasets}. The first four columns are from CLEVR, and the last four columns are from C{\footnotesize LEVR}T{\footnotesize EX}. $^\dag$ indicates post-processing. }
    \label{fig:clevr}
\end{figure*}

%% file: figs/movi_with_pp.tex
\begin{figure*}[t]
    \centering
 \rotatebox[origin=l]{90}{
 \parbox{0.09\linewidth}{ Our depth}
 \parbox{0.09\linewidth}{\centering Ours$^{\dag}$}
 \parbox{0.09\linewidth}{\centering Ours}
 \parbox{0.09\linewidth}{\centering PPMP$^{\dag}$}
 \parbox{0.09\linewidth}{\centering PPMP}
 \parbox{0.09\linewidth}{\centering GT Seg}
 \parbox{0.09\linewidth}{\centering Flow}
 \parbox{0.09\linewidth}{\centering RGB}}
 \hspace{0.001\linewidth} \includegraphics[width=0.093\linewidth]{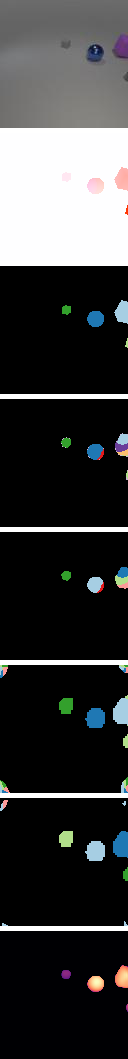}
    \includegraphics[width=0.093\linewidth]{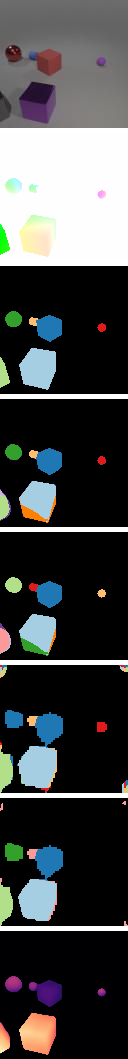}
    \hspace{0.002\linewidth}
    \includegraphics[width=0.093\linewidth]{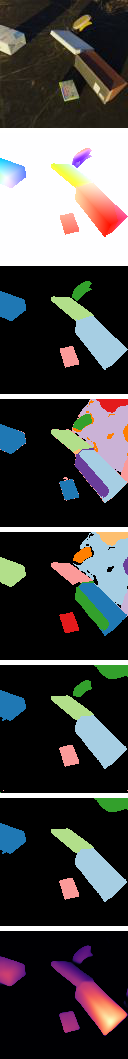}
    \includegraphics[width=0.093\linewidth]{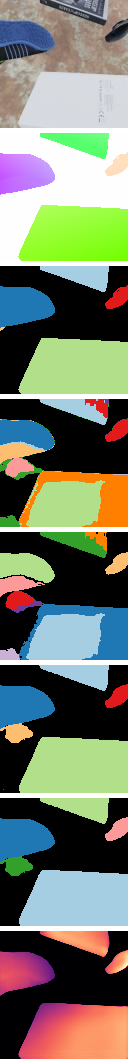}
    \hspace{0.002\linewidth}
    \includegraphics[width=0.093\linewidth]{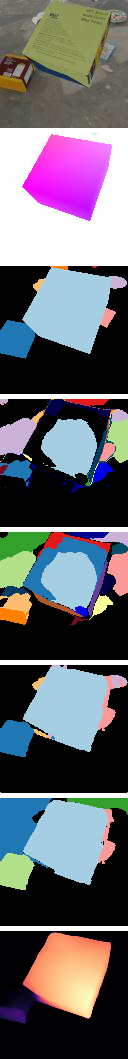}
    \includegraphics[width=0.093\linewidth]{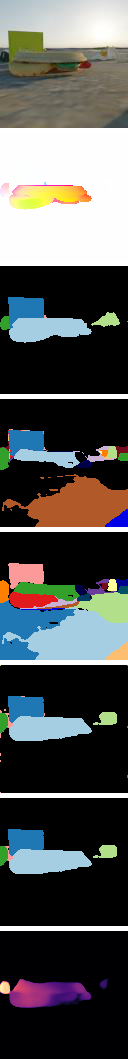}
    \hspace{0.002\linewidth}
    \includegraphics[width=0.093\linewidth]{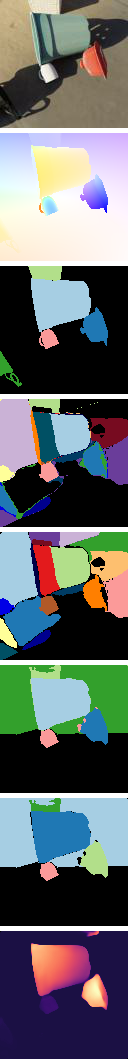}
    \includegraphics[width=0.093\linewidth]{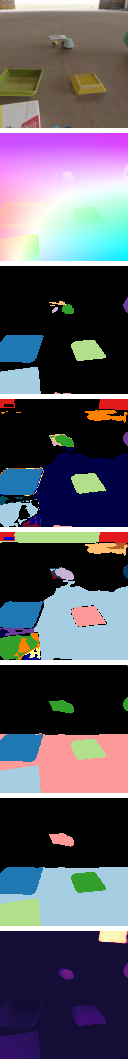}
        \\
     \quad MOVi-A  \quad \quad  \quad  \quad  \quad \quad MOVi-C \quad \quad \quad \quad  \quad  \quad  MOVi-D \quad \quad \quad \quad  \quad  \quad  \quad MOVi-E
    \caption{\textbf{Visualization of Depth and Segmentation Results on MOVi datasets}. $^\dag$ indicates post-processing. }
    \label{fig:movi-pp}
\end{figure*}

%% file: figs/kitti.tex
\begin{figure*}[b]
    \centering
     \rotatebox[origin=l]{90}{
 \parbox{0.07\linewidth}{\centering \footnotesize{RGB}}}
 \includegraphics[width=0.155\linewidth, height=0.09\linewidth]{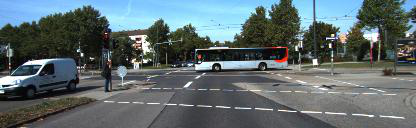}
 \includegraphics[width=0.155\linewidth, height=0.09\linewidth]{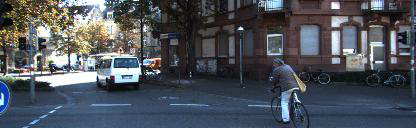}
 \includegraphics[width=0.155\linewidth, height=0.09\linewidth]{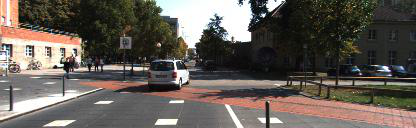}
 \includegraphics[width=0.155\linewidth, height=0.09\linewidth]{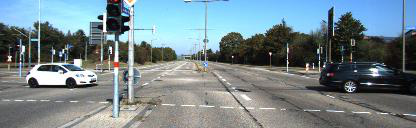}
\includegraphics[width=0.155\linewidth, height=0.09\linewidth]{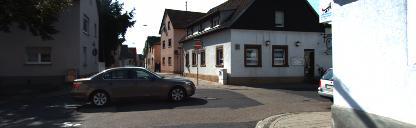}
 \includegraphics[width=0.155\linewidth, , height=0.09\linewidth]{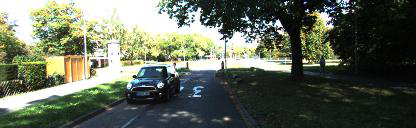}

 
 \rotatebox[origin=l]{90}{
 \parbox{0.07\linewidth}{\centering \footnotesize{our seg}}}
 \includegraphics[width=0.155\linewidth, height=0.09\linewidth]{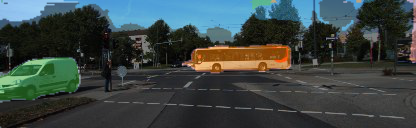}
 \includegraphics[width=0.155\linewidth, height=0.09\linewidth]{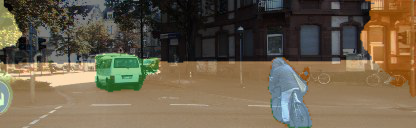}
 \includegraphics[width=0.155\linewidth, height=0.09\linewidth]{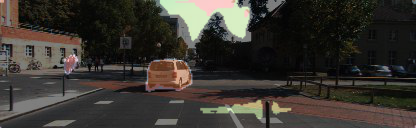}
 \includegraphics[width=0.155\linewidth, height=0.09\linewidth]{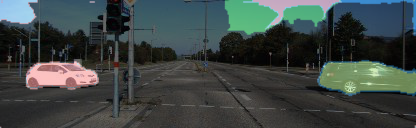}
\includegraphics[width=0.155\linewidth, height=0.09\linewidth]{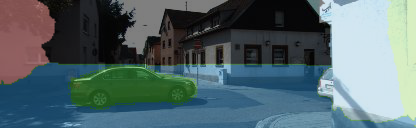}
 \includegraphics[width=0.155\linewidth,  height=0.09\linewidth]{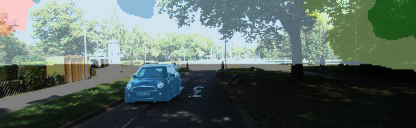}

 \rotatebox[origin=l]{90}{
 \parbox{0.07\linewidth}{\centering \footnotesize{our depth}}}
 \includegraphics[width=0.155\linewidth, height=0.09\linewidth]{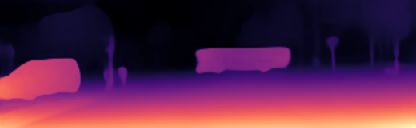}
 \includegraphics[width=0.155\linewidth, height=0.09\linewidth]{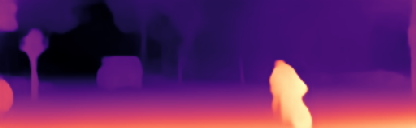}
 \includegraphics[width=0.155\linewidth, height=0.09\linewidth]{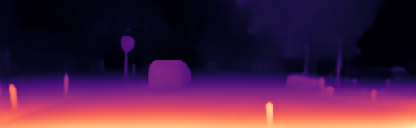}
 \includegraphics[width=0.155\linewidth, height=0.09\linewidth]{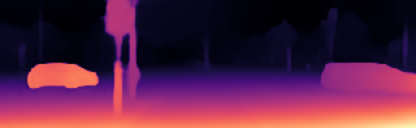}
\includegraphics[width=0.155\linewidth, height=0.09\linewidth]{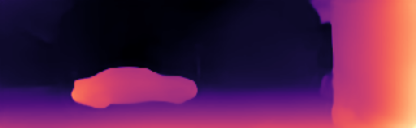}
 \includegraphics[width=0.155\linewidth, height=0.09\linewidth]{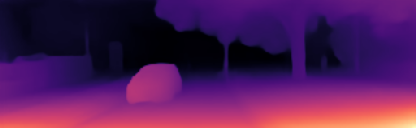}

    
    \caption{\textbf{Visualization of Segmentation and Depth Results on KITTI}.}
    \label{fig:kitti}
\end{figure*}

%% file: sec/supp/movidepth.tex
\section{Depth Evaluation on MOVi}
\label{sec:supp-movi-depth}
In this section, we evaluate the performance of our depth model on the foreground objects in each of the MOVi datasets. We evaluate the performance for both the Full model and the translation-only (Only-T) model. Note that as explained in the main paper, with the rotation-only model, the depth network is not trained. The results are presented in \tabref{tab:depth-movi}. We only evaluate the foreground objects. We use the median scaling approach~\cite{Zhou2017CVPR} to convert the predicted depths into the metric scale and cap the depths to 10~meters in all datasets.

The depth network of the Only-T models achieves better results on the MOVi\{C, D, E\} datasets. This is expected because the depth and segmentation networks are trained jointly. As a result, in the Full model, the errors in the estimation of rotation affect the depth estimations negatively, whereas, in the Only-T model, the depth estimations are not affected by erroneous rotation estimations. However, In the simpler MOVi-A dataset, as explained in the main paper, we found that the depth network in the Only-T model cannot predict the depth correctly. Therefore, we did not include the results of this version of the model for MOVi-A.
\input{tab/depth_movi}

%% file: tab/depth_movi.tex
\begin{table*}[]
\centering
\begin{tabular}{@{}lcccclccc@{}}
\toprule
Dataset & Abs Rel $\downarrow$ & Sq Rel $\downarrow$ & RSME $\downarrow$ & RMSE log $\downarrow$ & log10 $\downarrow$ & $\delta < 1.25$ $\uparrow$  & $ \delta < 1.25^2$ $\uparrow$ & $ \delta < 1.25^3$ $\uparrow$  \\ \midrule
MOVi-A (Full) & 0.113 & 0.348 & 1.483 & 0.226 & 0.061  & 0.813 & 0.912 & 0.949 \\
MOVi-A (Only-T) & - & - & - & - & -  & - & - & - \\
\midrule
MOVi-C (Full) & 0.225 & 0.604 & 1.845 & 0.299 & 0.100  & 0.609& 0.856 & 0.946 \\
MOVi-C (Only-T) & 0.166 & 0.446 & 1.437 & 0.217 & 0.068  & 0.779 & 0.932 & 0.978 \\
\midrule
MOVi-D (Full) & 0.544 & 2.863 & 3.744 & 1.381 & 0.415 & 0.348 & 0.547 & 0.657 \\
MOVi-D (Only-T) & 0.357 & 1.598 & 2.603 & 0.730 & 0.225 & 0.540 & 0.757 & 0.847 \\
\midrule
MOVi-E (Full) & 0.274 & 1.198 & 2.965 & 0.582 & 0.162 & 0.565 & 0.747 & 0.829 \\
MOVi-E (Only-T) & 0.244 & 0.989 & 2.596 & 0.559 & 0.159 & 0.596 & 0.761 & 0.842 \\
\bottomrule
\end{tabular}
\caption{\textbf{Depth Evaluation on Foreground Objects on MOVi Datasets}. Only-T refers to the version of our model where we only use the basis vectors corresponding to translation.}
\label{tab:depth-movi}
\end{table*}

%% file: sec/supp/kitti-intrinsics.tex
\section{KITTI with Intrinsics}
\label{sec:supp-intrinsics}
In our formulation, we produce basis vectors that do not depend on the values of focal lengths $f_x$ and $f_y$, which results in a set of 8 basis vectors as in \eqnref{eq:eight-basis}, instead of 6 as in \eqnref{eq:six-basis}, for each of $K$ regions. This means that our method can work without knowing the values of $f_x$ and $f_y$. On the other hand, monocular depth estimation methods assume a known intrinsics matrix, \ie $f_x, f_y, c_x,$ and $c_y$ are provided in the dataset. In order to make a fair comparison with monocular depth estimation methods, we train a version of our model on KITTI, where we also assume a known intrinsics matrix and create 6 basis vectors according to \eqnref{eq:six-basis} for each region using the values of known focal lengths. We report the depth estimation results with improved ground truth on the Eigen split of the KITTI dataset in \tabref{tab:kitti-intrinsics}. We can see that when we use a known camera intrinsic matrix (Ours-intrinsics), the performance is improved compared to our original model (Ours), and we can achieve better results that are comparable to the state-of-the-art in all metrics. 

\input{tab/kitti_intrinsics}

%% file: tab/kitti_intrinsics.tex
\begin{table*}[b]
\centering
\begin{tabular}{@{}lccccccc@{}}
\toprule
 & Abs Rel $\downarrow$ & Sq Rel $\downarrow$ & RSME $\downarrow$ & RMSE log $\downarrow$ & $\delta < 1.25$ $\uparrow$ & $ \delta < 1.25^2$ $\uparrow$ & $ \delta < 1.25^3$ $\uparrow$ \\ \midrule
Zhou \etal ~\cite{Zhou2017CVPR} & 0.176 & 1.532& 6.129 & 0.244 & 0.758 & 0.921 & 0.971 \\
Mahjourian \etal ~\cite{Mahjourian2018CVPR} & 0.134 & 0.983& 5.501 & 0.203 & 0.827 & 0.944 & 0.981 \\
GeoNet~\cite{Yin2018CVPR} & 0.132 & 0.994& 5.240 & 0.193 & 0.833 & 0.953 & 0.985 \\
DDVO~\cite{Wang2018CVPR} & 0.126 & 0.866& 4.932 & 0.185 & 0.851 & 0.958 & 0.986 \\
Ranjan \etal ~\cite{Ranjan2019CVPR} & 0.123 & 0.881& 4.834 & 0.181 & 0.860 & 0.959 & 0.985 \\
EPC++~\cite{Luo2019PAMI} & 0.120 & 0.789& 4.755 & 0.177 & 0.856 & 0.961 & 0.987 \\
Ours & 0.107 & 1.539& 4.027 & 0.149 & 0.911 & 0.971 & 0.989 \\
Monodepth2~\cite{Godard2019ICCV} & 0.090 & 0.545 & 3.942 & 0.137 & 0.914 & \underline{0.983} & \textbf{0.998} \\
Ours-intrinsics & \underline{0.084} & \underline{0.509} & \textbf{3.450} & \underline{0.132} & \underline{0.931} & 0.980 & 0.993 \\
PackNet-SfM~\cite{Guizilini2020CVPR} & \textbf{0.078} & \textbf{0.420} & \underline{3.485} & \textbf{0.121} & \textbf{0.934} & \textbf{0.986} & \underline{0.996} \\
\bottomrule
\end{tabular}
\caption{\textbf{Evaluation of Depth Estimation on KITTI.} We use the Eigen split of KITTI using improved ground truth. Note that all methods, except Ours, use the camera intrinsics matrix. Ours-intrinsics uses the intrinsics matrix and achieves comparable performance to state-of-the-art methods. }
\label{tab:kitti-intrinsics}
\end{table*}